\documentclass{article}

\usepackage[preprint]{neurips_2026}

\usepackage[utf8]{inputenc} 
\usepackage[T1]{fontenc}    
\usepackage{hyperref}       
\usepackage{url}            
\usepackage{booktabs}       
\usepackage{tabularx}       
\usepackage{array}          
\usepackage{amsfonts}       
\usepackage{nicefrac}       
\usepackage{microtype}      
\usepackage{xcolor}         
\usepackage{graphicx}       
\usepackage{subcaption}    
\usepackage{amsmath}        

\title{GRIP-VLM: Group-Relative Importance Pruning for Efficient Vision-Language Models}

\author{%
  Mingzhe Huang\textsuperscript{1} \quad 
  Weijun Wang\textsuperscript{1,}\thanks{Corresponding authors.} \quad 
  Xin Ding\textsuperscript{1,4} \quad 
  Liang Mi\textsuperscript{1,3} \quad 
  Hao Wen\textsuperscript{1} \\
  \textbf{Yuanchun Li\textsuperscript{1} \quad 
  Lichen Pang\textsuperscript{2} \quad 
  Shansong Yang\textsuperscript{2} \quad 
  Yunxin Liu\textsuperscript{1} \quad 
  Ting Cao\textsuperscript{1,}\protect\footnotemark[1]} \\
  \\
  \textsuperscript{1}Institute for AI Industry Research (AIR), Tsinghua University \\
  \textsuperscript{2}Juhaokan Technology Co.,Ltd \\
  \textsuperscript{3}Nanjing University \quad 
  \textsuperscript{4}University of Science and Technology of China
}
\usepackage{enumitem}
\begin{document}

\maketitle

\begin{abstract}
  In Vision-Language Models (VLMs), processing a massive number of visual tokens incurs prohibitive computational overhead. While recent training-aware pruning methods attempt to selectively discard redundant tokens, they largely rely on continuous-gradient relaxations. However, visual token pruning is inherently a discrete, non-convex combinatorial problem; consequently, these continuous approximations frequently trap the optimization in sub-optimal local minima, especially under aggressive compression budgets. To overcome this fundamental bottleneck, we propose GRIP-VLM, a Group-Relative Importance Pruning framework driven by Reinforcement Learning. Rather than relying on smooth-gradient assumptions, GRIP-VLM formulates pruning as a Markov Decision Process, employing a Group Relative Policy Optimization (GRPO) paradigm anchored by supervised warm-up to directly explore the discrete selection space. Integrated with a budget-aware scorer, our lightweight agent dynamically evaluates per-token importance and adapts to arbitrary compression ratios without retraining. Extensive experiments across diverse multimodal benchmarks demonstrate that GRIP-VLM consistently outperforms heuristic and supervised-learning baselines, achieving a superior Pareto frontier and delivering up to a 15\% inference speedup at equal accuracy.
\end{abstract}
\section{Introduction}
\label{sec:intro}

Visual-Language Models (VLMs)~\cite{li2023blip, zhu2023minigpt, liu2024improved, liu2024visual} extend large language models (LLMs)~\citep{touvron2023llama, grattafiori2024llama, bai2023qwen, yang2024qwen2, wang2025data} with the capability to process and understand visual information, enabling strong performance on tasks such as image captioning, visual question answering (VQA), video understanding~\citep{wang2024internvideo2}, and multi-modal reasoning~\citep{wang2024exploring}.
Unlike LLMs, which process only a small number of information-dense text tokens, VLMs introduce a large number of visual tokens, ranging from 576 in early models~\citep{liu2023visualllava} to over 16,384 in recent systems~\citep{bai2025qwen25vltechnicalreport}, which creates significant computational overhead and can be further exacerbated by high-resolution images~\citep{li2024mini} and multi-frame video inputs~\citep{tang2023video}.

Token pruning has emerged as a critical research direction to address this bottleneck.
Training-free heuristic methods identify and discard redundant visual tokens based on handcrafted criteria: sampling-based approaches~\citep{shang2024llavaprumerge} adaptively pool or subsample tokens; attention-based methods~\citep{chen2024imagefastv, zhang2025sparsevlm} prune tokens with low cross-modal attention scores; and similarity-based methods~\citep{bolya2023tokenmergingvitfaster, fu2025framefusioncombiningsimilarityimportance} merge tokens with high feature similarity.
While computationally efficient, these methods rely on fixed, model-specific heuristics, making them brittle under distribution shift and incapable of adapting to per-sample input complexity.

\begin{figure}[htbp]
    \vspace{-1em}
    \centering
    \begin{minipage}[t]{0.48\linewidth}
        \centering
        \includegraphics[width=\linewidth]{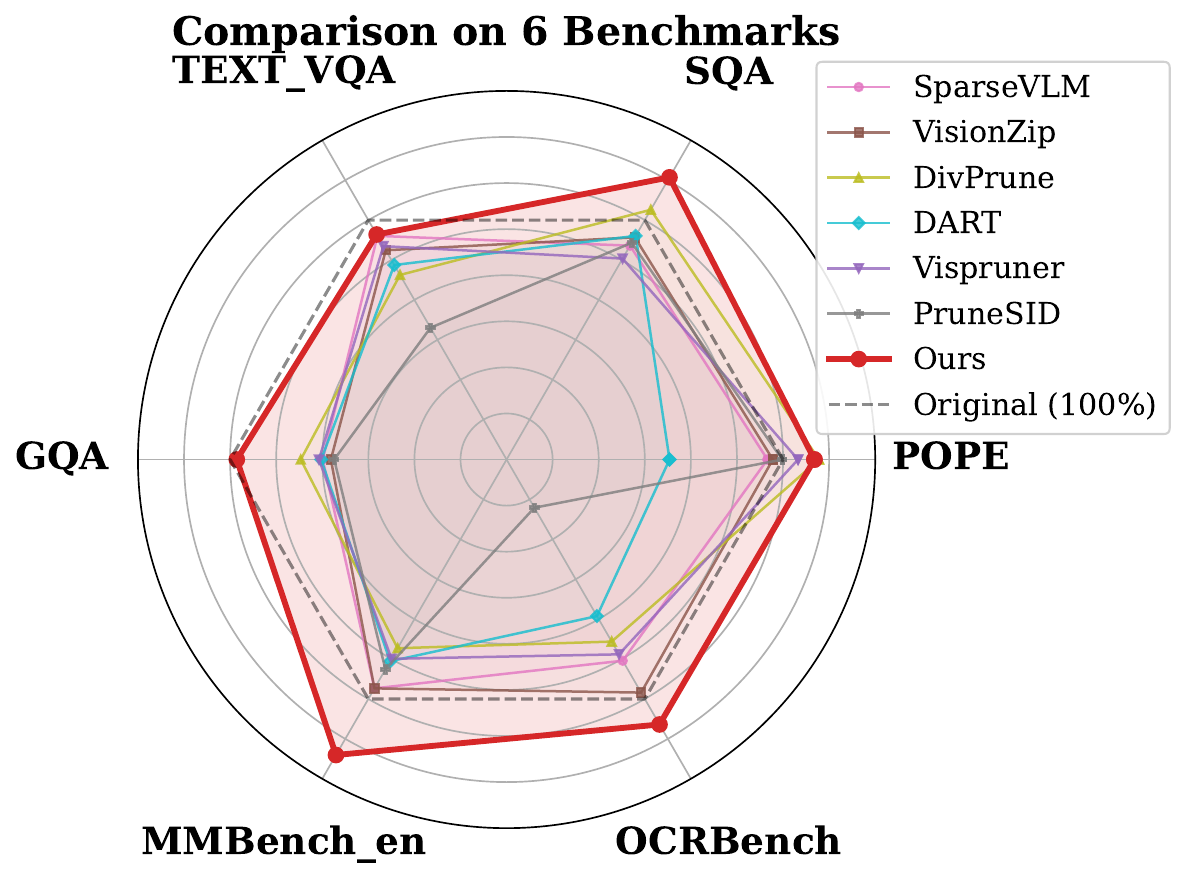}
        \caption{Radar chart comparing GRIP-VLM against baseline methods.}
        \label{fig:radar}
    \end{minipage}
    \hfill
    \begin{minipage}[t]{0.48\linewidth}
        \centering
        \includegraphics[width=\linewidth]{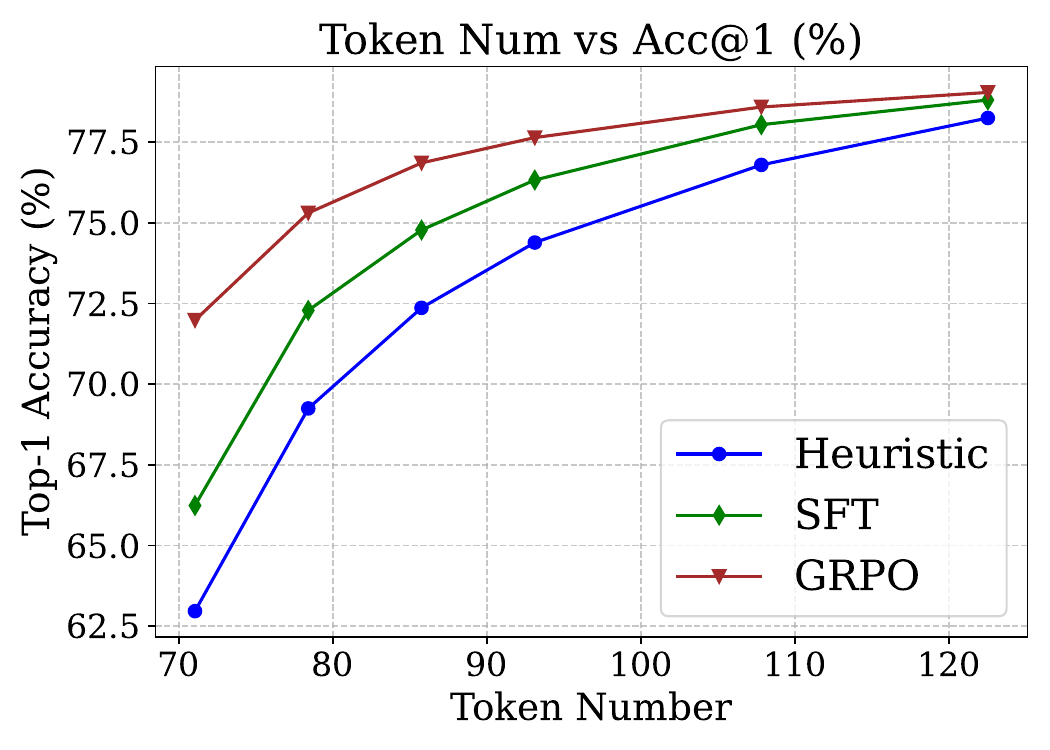}
        \caption{Top-1 accuracy of heuristic, SL, and RL pruning applied to ViT on ImageNet.}
        \label{fig:acc1}
    \end{minipage}
    \vspace{-1em}
\end{figure}

To overcome the rigidity of heuristics, recent training-aware methods~\citep{rao2021dynamicvitefficientvisiontransformers, wang2024smarttrimadaptivetokensattention, zhu2025visionselectorendtoendlearnablevisual} train a learned scorer to estimate per-token importance, using techniques such as Gumbel-Softmax or differentiable top-$K$ to enable gradient-based optimization through discrete pruning decisions.
However, we argue that these methods share a common fundamental flaw: token pruning is inherently a \textbf{combinatorial subset-selection problem}, that preserves $K$ visual tokens to minimize the loss of VLM generation quality $f$ of given inputs $x$, with a discrete and non-convex decision space,
\begin{equation}
\min_{\mathbf{z} \in \{0,1\}^N} \; \mathcal{L}\bigl(f(x;\mathbf{z})\bigr) \quad \text{s.t.} \quad \textstyle\sum_{i}z_i \le K,
\end{equation}
yet all continuous-relaxation approaches implicitly assume a smooth, convex surrogate landscape.
As evidenced by EPIC~\citep{wen2025efficientmultimodallargelanguage}, this mismatch causes optimization to converge to suboptimal local minima, particularly under aggressive compression.

To bridge the gap between the combinatorial nature of pruning and the limitations of current optimization techniques, we propose \textbf{GRIP-VLM}, a two-stage dynamic pruning framework designed for efficient VLM inference.
Rather than relying solely on continuous relaxation, we synergize curriculum learning with reinforcement learning (RL) to effectively navigate the complex, non-convex pruning decision space.
Our core contributions are threefold:

\begin{itemize}[leftmargin=*]
    \item \textbf{System Dimension: A Universal and Compatible Framework.}
GRIP-VLM integrates lightweight scoring modules into Transformer layers with negligible overhead, supporting both the visual encoder and LLM decoder stages.
It is orthogonal to existing heuristic methods and can be combined with them for further gains.

    \item \textbf{Token-wise Dimension: A Budget-Aware Scorer.}
We design a scorer that explicitly perceives the current pruning pressure via a budget-aware modulation mechanism, trained with curriculum learning to progressively adapt from lossless to high-compression regimes.
This enables robust generalization across arbitrary token budgets without retuning.

    \item \textbf{Optimization Dimension: A Hybrid RL Paradigm.}
We propose a SFT-anchored exploration strategy: supervised distillation initializes a strong baseline policy, then GRPO-based RL explores the discrete selection space to escape local optima and maximize task reward,without relying on the smooth-gradient assumption that limits SL methods.
\end{itemize}

Extensive experiments across diverse multi-modal benchmarks demonstrate that GRIP-VLM effectively reduces the computational overhead of VLMs without sacrificing performance.
As shown in Figure~\ref{fig:radar}, GRIP-VLM achieves the best overall performance across 6 benchmarks at a budget of 192 tokens.
Figure~\ref{fig:acc1} further shows that on ViT\citep{dosovitskiy2021imageworth16x16words}, heuristic, Supervised Learning(SL), and Reinforcement Learning(RL) pruning yield progressively higher accuracy, validating the benefit of each stage in our framework.
Under an equal-accuracy constraint, our method achieves \textbf{15\%} faster inference than competing approaches.
On hard samples where conventional pruning methods struggle , GRIP-VLM improves the success rate by \textbf{33\%} on average over SOTA baselines.

\section{Related Work}
\label{sec:related}

\paragraph{High-Resolution VLMs and Efficiency Bottleneck.}
Recent advances in VLMs~\citep{liu2023visualllava, zhu2023minigpt, instructblip} have achieved remarkable fine-grained perception via dynamic or arbitrary-resolution strategies~\citep{liu2024llavanext, li2024llava, wang2024qwen2, bai2025qwen2}. 
However, scaling visual tokens to preserve high-frequency details (\emph{e.g.}, up to 16,384 tokens in Qwen2.5-VL~\citep{bai2025qwen25vltechnicalreport}) introduces prohibitive computational and memory costs due to the quadratic complexity of self-attention~\citep{vaswani2017attention}. 
This exacerbates the need for efficient visual token compression.

\paragraph{Training-Free Heuristics.}
Existing compression techniques largely rely on training-free heuristics. \textit{Sampling-based methods}~\citep{shang2024llavaprumerge} use spatial pooling and outlier detection but suffer from inflexible, predefined compression ratios. \textit{Attention-based strategies}~\citep{chen2024imagefastv, zhang2025sparsevlm} prune tokens with low cross-modal attention scores, though they remain vulnerable to attention sinks and semantic drift~\citep{zhang2024beyond}. Alternatively, \textit{similarity-based methods} like Token Merging~\citep{bolya2023tokenmergingvitfaster} and FrameFusion~\citep{fu2025framefusioncombiningsimilarityimportance} aggregate adjacent tokens, which inevitably discards subtle but critical local details. Inherently biased towards contiguous visual ``blobs'' and specific feature distributions, these heuristics struggle to generalize across diverse input complexities.

\paragraph{Learnable Token Selection.}
To overcome heuristic rigidity, training-aware methods learn to select tokens dynamically. Approaches like DynamicViT~\citep{rao2021dynamicvitefficientvisiontransformers} and SmartTrim~\citep{wang2024smarttrimadaptivetokensattention} employ the Gumbel-Softmax trick, while VisionSelector~\citep{zhu2025visionselectorendtoendlearnablevisual} utilizes differentiable Top-K formulations, often guided by knowledge distillation. However, a fundamental bottleneck remains: applying continuous, smooth-gradient relaxations to the inherently non-convex, discrete subset selection problem often traps the policy in sub-optimal local minima, especially under aggressive compression budgets. Unlike these approaches, GRIP-VLM completely bypasses continuous relaxations, leveraging Reinforcement Learning to directly explore the discrete combinatorial space for optimal, fine-grained token retention.

\section{Motivation: Heterogeneity of the Token Pruning Landscape}
\label{sec:motivation}

\begin{figure*}[htbp]
  \centering
  \begin{minipage}[t]{0.49\linewidth}
    \centering
    \includegraphics[width=\linewidth]{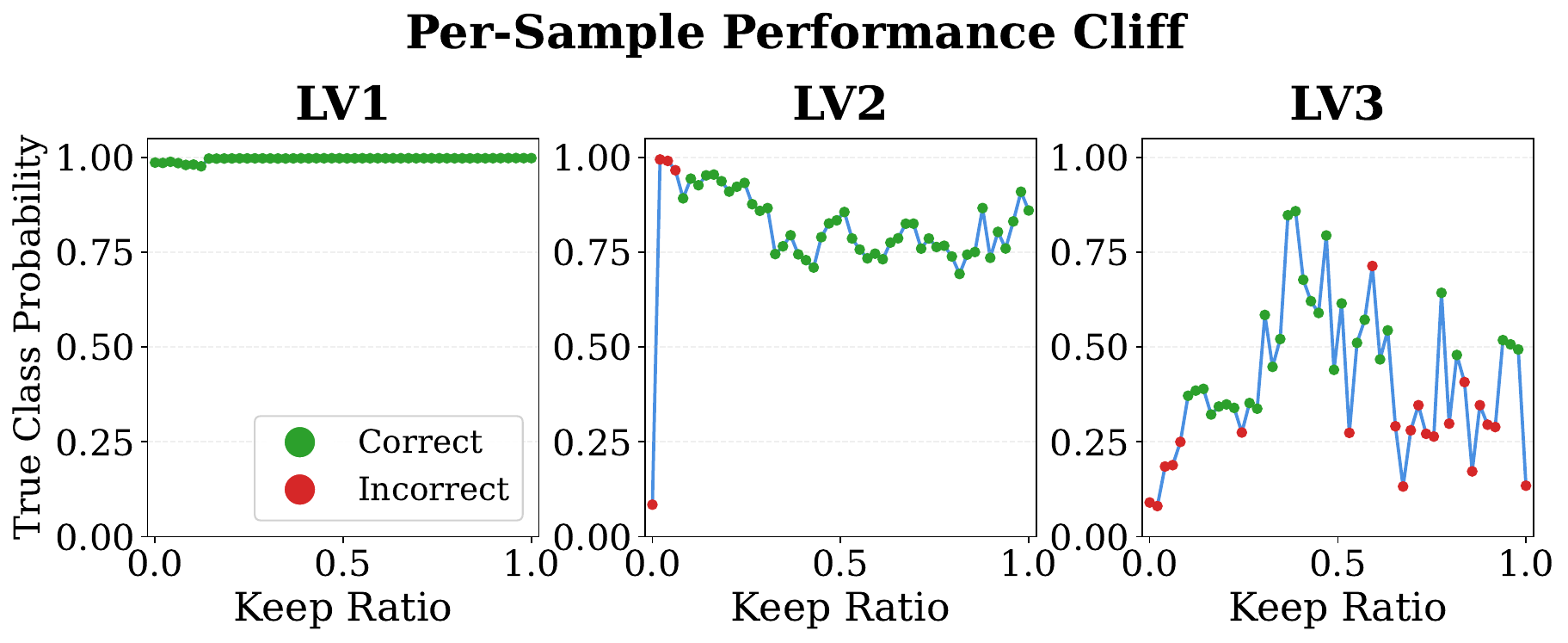}
  \end{minipage}
  \hfill
  \begin{minipage}[t]{0.49\linewidth}
    \centering
    \includegraphics[width=\linewidth]{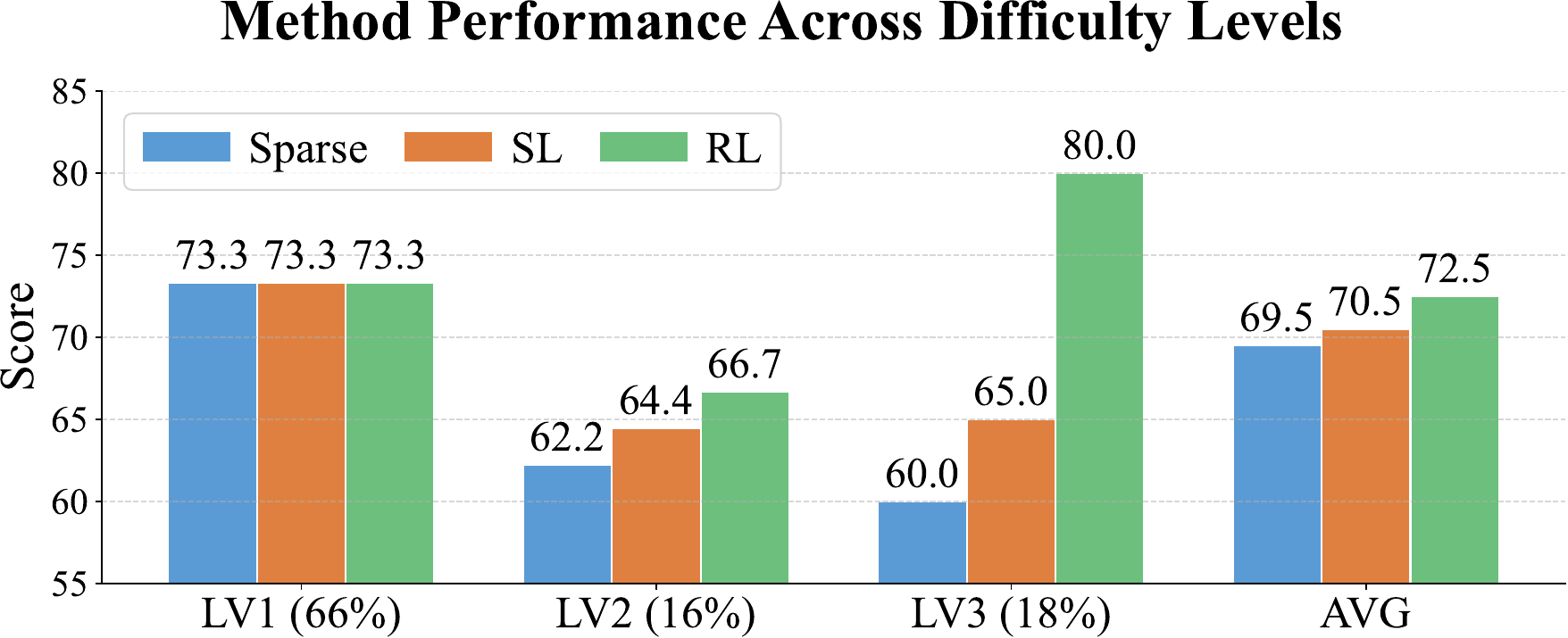}
  \end{minipage}
  \caption{\textbf{Left: Per-Sample Performance Cliff (SQA).} Each curve shows the true class probability as the keep ratio varies; green/red dots indicate correct/incorrect predictions.
  \textbf{Right: Method Performance Across Difficulty Levels (SQA).} Accuracy of sparse (heuristic), SL, and RL across difficulty levels (LV1--LV3) and overall average. 
  Complete results are provided in Appendix~\ref{app:full_motivation}.}
  \label{fig:cliff_and_levels}
\vspace{-0.5em}
\end{figure*}

A key observation motivating our design is that not all pruning tasks are equally challenging.
To characterize this heterogeneity, we apply a heuristic pruning method to each sample across a range of keep ratios from 0 to 1, and count the number of  alternations between correct (green) and incorrect (red) predictions as the keep ratio varies.
A transition count of 0 indicates that pruning has virtually no effect on the prediction, yielding a flat, robust curve; a count of 1 indicates a roughly monotone degradation; and a count of $\geq 2$ indicates a highly oscillatory curve where correct and incorrect predictions alternate repeatedly.
We accordingly define three difficulty levels in Figure\ref{fig:cliff_and_levels} (left):

\begin{itemize}[leftmargin=*]
  \item \textbf{LV1 (Easy).} The prediction is stable across all keep ratios. These samples are highly robust to pruning, and a simple heuristic scorer already achieves near-optimal performance. 
  \item \textbf{LV2 (Medium).} The success rate decreases roughly monotonically as the keep ratio drops. The landscape is well-structured, an SL-trained scorer can reliably improve over the heuristic baseline.
  \item \textbf{LV3 (Hard).} The curve is highly non-monotone, with correct and incorrect predictions alternating as the keep ratio changes. Such samples are difficult for heuristic methods to handle and present a challenging optimization target for SL. Only RL, through extensive exploration and trial-and-error, can effectively navigate this irregular landscape.
\end{itemize}

To quantify the distribution of difficulty levels in practice, we randomly sampled 200 images from each of three benchmarks (MMBench, POPE, SQA), applied the heuristic scorer to classify each sample, and evaluated all three methods (sparse, SL, RL) per level.
Figure~\ref{fig:cliff_and_levels} (right) reports the results.
As expected, all three methods perform comparably on LV1 samples.
On LV2, SL already yields a clear improvement over the heuristic baseline.
On LV3, RL achieves the largest gains,up to 20 points over the heuristic and SL baselines,confirming that RL's exploratory training is essential for hard, non-monotone samples.

\subsection{Mathematical Pathologies of LV3 and the Necessity of Token-Level Credit Assignment}
\label{subsec:lv3_analysis}

\textbf{Problem Setup.} Let $\mathcal{T}$ be the set of $N$ visual tokens, $\pi \subseteq \mathcal{T}$ a retention mask, and $P(\pi)$ the task performance (or $\mathrm{correct}(\pi) \in \{0,1\}$). LV3 samples exhibit severe loss landscapes that violate standard optimization assumptions:

\begin{itemize}[leftmargin=*]
    \setlength{\itemsep}{1pt}
    \setlength{\parskip}{0pt}
    \setlength{\parsep}{0pt}
    \item \textbf{(P1) Non-monotonicity}: Keeping more tokens intuitively preserves performance. LV3 violates this: $\exists \pi_1 \subset \pi_2$ such that $P(\pi_1) > P(\pi_2)$.
    \item \textbf{(P2) Non-Lipschitz slope}: Removing a single critical token $t^*$ can flip the prediction entirely, i.e., $|P(\pi) - P(\pi \setminus \{t^*\})| \approx 1$. The local gradient diverges, trapping continuous relaxations.
    \item \textbf{(P3) Non-additive importance}: Token contributions are coupled (e.g., XOR structure). For misleading tokens $A, B$, we may have $\mathrm{correct}(\mathcal{T} \setminus \{A\}) = 0$ and $\mathrm{correct}(\mathcal{T} \setminus \{B\}) = 0$, but $\mathrm{correct}(\mathcal{T} \setminus \{A,B\}) = 1$. Independent scoring fails here.
\end{itemize}

\textbf{Credit assignment dilemma.} To navigate these pathologies, we compare two learning objectives for mask optimization:
\begin{align}
    \mathcal{L}_{\mathrm{SFT}} &\propto -R(\pi)\sum_{t \in \pi} \log p_t - (1-R(\pi))\sum_{t \notin \pi} \log(1-p_t), \label{eq:sft_loss} \\
    \nabla_{s_t} J_{\mathrm{RL}} &\approx \frac{1}{G}\sum_{j=1}^{G} \big(R(\pi^{(j)}) - \bar{R}\big)\big(a_t^{(j)} - p_t\big), \label{eq:rl_loss}
\end{align}
where $s_t$ is the log-odds score, $p_t = \sigma(s_t)$, and $a_t^{(j)} \in \{0,1\}$ is the inclusion indicator in rollout $j$.

\begin{figure}[t]
  \centering
  \includegraphics[width=0.8\linewidth]{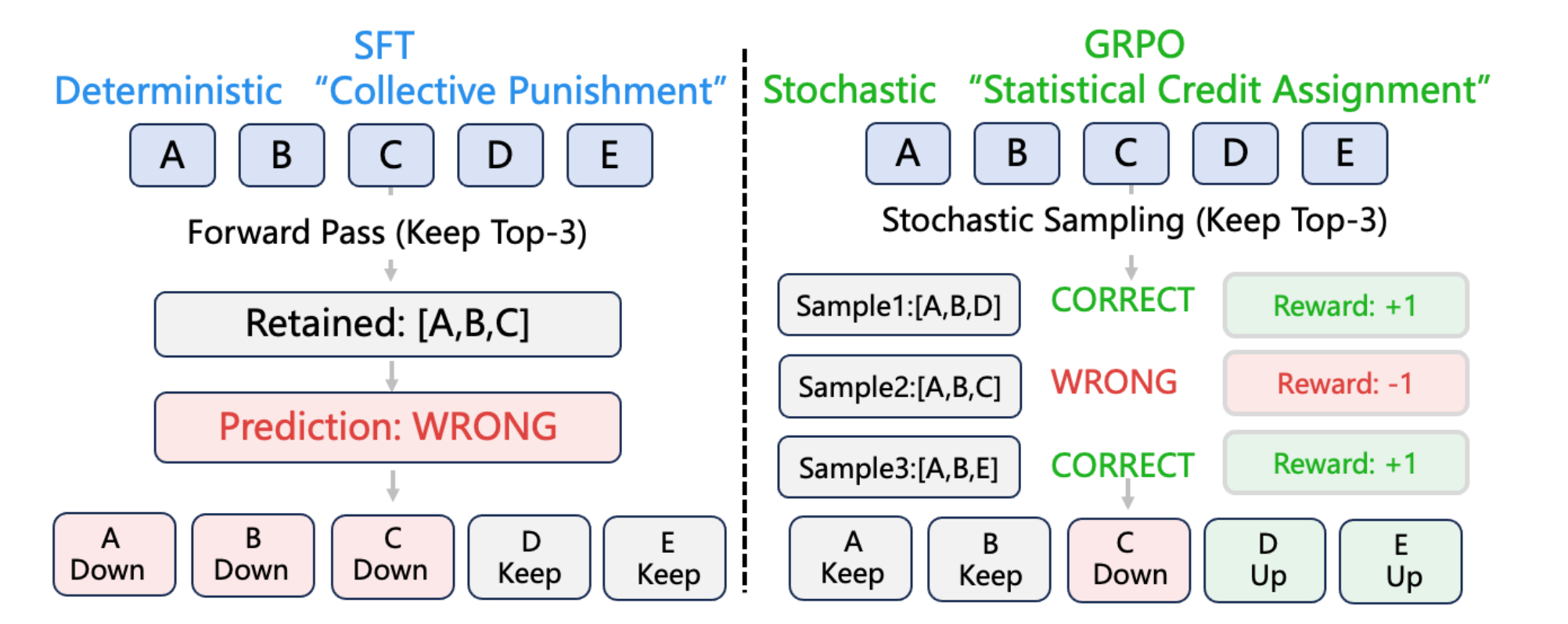}
  \caption{\textbf{Credit Assignment: SFT vs.\ GRPO.} SFT uniformly penalizes all tokens in a failing mask (left), while GRPO isolates the true culprit via stochastic rollouts (right).}  
  \label{fig:credit_assignment}
  \vspace{-1em}
\end{figure}
\begin{itemize}
[leftmargin=*]
    \setlength{\itemsep}{1pt}
    \setlength{\parskip}{0pt}
    \setlength{\parsep}{0pt}
    \item \textit{SFT (Set-level)}: Applies a uniform penalty to all $t \in \pi$ when $R(\pi)=0$. If $\pi=\{A,B,C\}$ fails purely due to $C$, useful tokens $A$ and $B$ are wrongfully penalized (Figure~\ref{fig:credit_assignment}, left).
    \item \textit{GRPO (Token-level)}: Isolates token contributions via $G$ stochastic rollouts. Tokens appearing equally in winning and losing trajectories ($A, B$) receive zero net penalty, while the true culprit $C$ incurs a negative update (Figure~\ref{fig:credit_assignment}, right).
\end{itemize}

\textbf{Two-stage necessity.} Since no single objective optimally handles all regimes, GRIP-VLM employs SL warm-up to rapidly solve LV1/LV2 (where (P1)-(P3) are mild), followed by RL exploration to navigate the LV3 pathologies.

\section{Method}
\label{sec:method}

\begin{figure*}[t]
  \centering
  \includegraphics[width=0.9\textwidth]{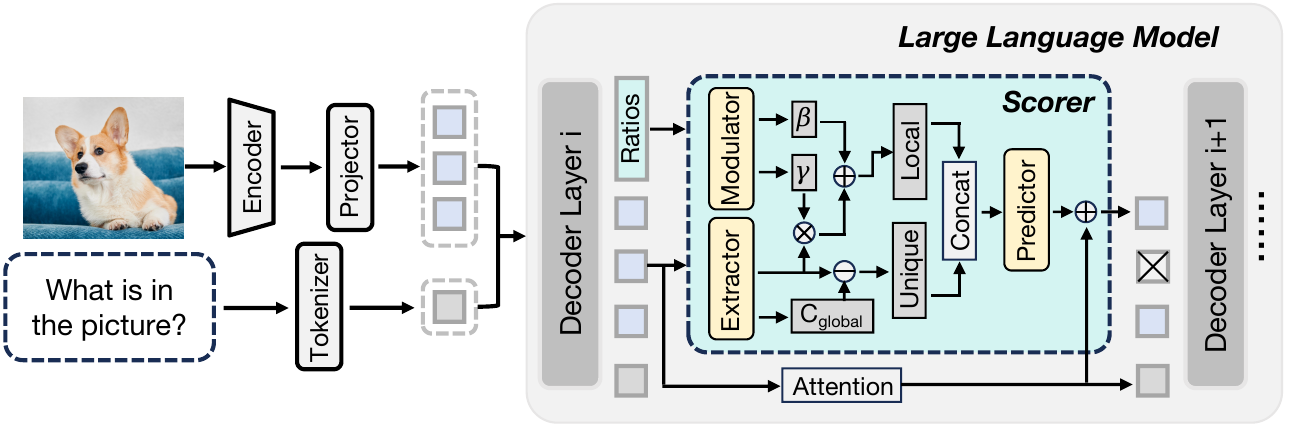}
  \caption{Inference architecture of GRIP-VLM. At each pruning stage within the LLM decoder, a lightweight Scorer takes the token representations, current compression ratio, and the target compression ratio as input, and outputs binary keep/drop decisions for each visual token. }
  \label{fig:infra}
  \vspace{-1.5em}
\end{figure*}

We present GRIP-VLM in Figure~\ref{fig:infra}, a hierarchical dynamic pruning framework designed to progressively mitigate visual redundancy while preserving essential semantic information.
Instead of imposing pruning constraints at every transformer layer, which could introduce non-negligible computational overhead, our framework is strategically interspersed at specified intervals within the VLM backbone.
At each designated pruning stage, GRIP-VLM relies on a unified \textbf{Scorer} module as its core decision-maker.
The Scorer performs fine-grained, contextualized token-wise importance evaluation and facilitates the selection of critical visual features, ensuring an optimal balance between computational efficiency and model performance.

\subsection{Adaptive Token Scorer Architecture}
\label{subsec:adaptive_scorer}

Our Adaptive Token Scorer (ATS) evaluates token importance by integrating local semantics, global redundancy, and dynamic pruning budgets into a unified scoring mechanism. We summarize the three core components of the ATS below:

\paragraph{Budget-Aware Environmental Modulation.}
The optimal token retention strategy inherently shifts with the available compression budget. To enable a single scorer to handle arbitrary pruning rates without retraining, we inject dynamic pruning pressure into the feature space. Specifically, we encode the gap between the target pruning ratio ($\rho_{\mathrm{target}}$) and the current ratio ($\rho_{\mathrm{now}}$), alongside image-level context, to predict affine parameters $[\gamma, \beta]$ via a FiLM layer \cite{perez2017filmvisualreasoninggeneral}. The local features are then dynamically modulated:
\begin{equation}
    \tilde{F}_{\mathrm{local}} = F_{\mathrm{local}} \odot (1 + \gamma) + \beta.
\end{equation}
This environmental modulation allows the scorer to adaptively relax or tighten its selection criteria based on real-time budget constraints.

\paragraph{Comprehensive Feature Scope (Local + Global).}
A token’s intrinsic value depends on both its local semantic richness (e.g., salient foreground objects) and its uniqueness relative to other retained tokens. For an input token $x_i$, we extract a local representation $F_{\mathrm{local}}$ via a lightweight extractor. To explicitly penalize redundancy, we aggregate the currently retained tokens via masked pooling to form a global context summary $C_{\mathrm{global}}$. The feature’s uniqueness is then elegantly captured by its deviation from this global context: $F_{\mathrm{unique}} = F_{\mathrm{local}} - C_{\mathrm{global}}$.

\paragraph{Adaptive Heuristic Fusion.}
To ensure optimization stability and leverage established prior knowledge (e.g., attention weights or feature norms, denoted as $S_{\mathrm{heuristic}}$), we introduce a dynamic gated fusion mechanism. Using the concatenated feature $[\tilde{F}_{\mathrm{local}}, F_{\mathrm{unique}}]$, the ATS predicts both a learned score $S_{\mathrm{ours}}$ and a dynamic fusion weight $\alpha \in (0, 1)$. The final token retention score is formulated as a convex combination:
\begin{equation}
    S_{\mathrm{final}} = \alpha \cdot S_{\mathrm{ours}} + (1 - \alpha) \cdot S_{\mathrm{heuristic}}.
\end{equation}
This adaptive gating provides a safe fallback to heuristics during early training or extreme compression regimes, while allowing the model to progressively rely on its learned representations as the policy matures.

\subsection{Optimization and Training Strategy}

Training the pruning policy end-to-end from scratch is unstable: RL exploration is sample-inefficient without a reasonable starting point, while pure supervised distillation cannot escape the local optima imposed by the smooth-gradient assumption.
We therefore adopt a two-stage paradigm that first builds a strong initialization via supervised distillation, then refines it through RL exploration (see Appendix~\ref{app:train_pipeline} for an illustration).
Stage~I gives the scorer a reliable sense of token importance under the teacher's guidance; Stage~II then uses this warm-start policy to explore the combinatorial selection space efficiently, discovering token subsets that the distillation objective alone cannot reach.

\subsubsection{Stage I: Scorer Distillation and Warm-up.}
In the first stage, we focus on training the \textbf{Adaptive Token Scorer (ATS)} to produce robust importance estimations. We utilize a teacher-student distillation framework where the full-token model acts as the teacher. The Scorer is optimized via a multi-task loss function:
\begin{equation}
    \mathcal{L}_{\mathrm{scorer}} = \mathcal{L}_{\mathrm{distill}} + \alpha\mathcal{L}_{\mathrm{KL}} + \beta\mathcal{L}_{\mathrm{ratio}},
\end{equation}
where $\mathcal{L}_{\mathrm{distill}}$ is the feature-based distillation loss (e.g., MSE) that aligns the intermediate feature maps of the pruned student with the teacher, $\mathcal{L}_{\mathrm{KL}}$ minimizes the Kullback-Leibler divergence between their output logits, and $\mathcal{L}_{\mathrm{ratio}}$ applies a sparsity constraint to prevent trivial dense solutions.

To improve generalization, we introduce a curriculum learning strategy over the equivalent token count $T_{\mathrm{eq}} = \sum_{i=1}^K r_i \cdot N_0 \cdot L_i$, where $N_0$ is the initial token capacity and $L_i$ the number of layers at stage $i$. We gradually compress $T_{\mathrm{eq}}$ during training, enabling the scorer to learn robust pruning patterns across diverse compression rates.

\subsubsection{Stage II: Policy Optimization via GRPO}

To escape the local minima inherent in continuous relaxations, we freeze the VLM backbone and optimize the \textbf{Scorer} as an RL policy $\pi_\theta$. Given intermediate feature maps $s$, the Scorer outputs retention probabilities $p_i$, from which discrete pruning decisions are independently sampled: $a_i \sim \text{Bernoulli}(p_i)$.

We employ GRPO to optimize $\pi_\theta$, circumventing the instability of training an auxiliary critic network in highly volatile reward landscapes. For each input $s$, we sample a group of $G$ independent pruning trajectories $\{a^{(1)}, \dots, a^{(G)}\}$. The policy is updated by maximizing the GRPO objective:
\begin{equation}
    \mathcal{J}_{\mathrm{GRPO}}(\theta) = \mathbb{E} \left[ \frac{1}{G} \sum_{j=1}^G \left\{ \min \Big( \rho_j \hat{A}_j, \, \text{clip}(\rho_j, 1-\epsilon, 1+\epsilon)\hat{A}_j \Big) - \beta \mathbb{D}_{\mathrm{KL}}[\pi_\theta \,\|\, \pi_{\mathrm{ref}}] \right\} \right],
\end{equation}
where $\rho_j = \pi_\theta(a^{(j)}|s) / \pi_{\theta_{\mathrm{old}}}(a^{(j)}|s)$ is the probability ratio. The group-relative advantage $\hat{A}_j = (r_j - \bar{r})/\sigma(r)$ is computed from the composite trajectory reward $r_j$, which explicitly balances task performance against the computational budget constraint:
\begin{equation}
    r_j = R_{\mathrm{task}}^{(j)} - \gamma \cdot \max\!\big(0,\, T_{\mathrm{eq}}^{(j)} - T_{\mathrm{target}}\big).
\end{equation}

\paragraph{Hybrid Task Reward Design.}
A principled definition of $R_{\mathrm{task}}$ is critical and must inherently adapt to the underlying task format~\cite{chen2025sftrlearlyinvestigation}. We unify heterogeneous tasks via a dynamically scaled reward:

\begin{itemize}[leftmargin=*]
    \setlength{\itemsep}{2pt}
    \setlength{\parskip}{0pt}
    \item \textbf{Verifiable tasks} (e.g., multiple-choice, binary judgment): We adopt a sparse binary correctness reward, where $R_{\mathrm{task}} = 1$ if $\hat{y} = y$, and $0$ otherwise. This unambiguous discrete signal is strictly faithful to task quality and immune to reward hacking via logit recalibration.

    \item \textbf{Open-ended tasks} (e.g., free-form generation): Since exact-match evaluation is ill-posed here, we anchor the reward to the Stage I SFT model. Using the SFT model's target loss $L_{\mathrm{ref}}$ on the same sample as a baseline, the pruned model's loss $L$ is mapped to a normalized score:
    \begin{equation}
        R_{\mathrm{task}} = 1 - \sigma\!\big(\alpha \cdot (L - L_{\mathrm{ref}})\big).
    \end{equation}
    This scaled transformation achieves two vital goals. First, it normalizes the continuous loss into $[0, 1]$, bridging the scale gap with verifiable tasks to stabilize GRPO advantage estimation. Second, it naturally instantiates our \emph{SFT-anchored exploration} paradigm: the agent earns positive incentives ($R_{\mathrm{task}} > 0.5$) strictly by discovering token combinations that surpass the SFT initialization. 
\end{itemize}

\section{Experiments}
\label{sec:experiments}

\subsection{Experimental Setup}

\paragraph{Evaluation.}
We evaluate on 8 image-based benchmarks across 2 VLMs (LLaVA-1.5-7B and LLaVA-1.5-13B), applying GRIP-VLM at both the visual encoder and LLM decoder stages to accommodate different inference scenarios.
Baseline and benchmark details are provided in the Appendix.

\paragraph{Implementation Details.}
We build on LLaVA-1.5-7B and LLaVA-1.5-13B as backbone models and train on the LLaVA-665K instruction-tuning dataset.
The Decoder Scorer is inserted at layers 3, 7, and 16 of the LLM decoder, performing $K{=}3$ progressive pruning stages.The Encoder Scorer is inserted between the Vision Encoder and the Projector, performing a single pruning stage $K{=}1$.
All LLM parameters are frozen throughout training; only the Scorer is optimized with Adam at a learning rate of $1{\times}10^{-5}$ for 10 epochs.
In Stage~II, GRPO samples $G{=}16$ rollouts per instance to compute group-relative advantages.

\subsection{Main Results}

Table~\ref{tab:results} reports accuracy across all benchmarks under three token budgets: 192, 128, and 64 tokens.

\begin{table*}[htbp]
    \vspace{-1em}
    \centering
    \caption{Results under different experiments.}
    \label{tab:results}
    \resizebox{\textwidth}{!}{%
    \begin{tabular}{lrrrrrrrrr}
    \toprule
        llava-v1.5-7b & \textbf{MME} & \textbf{POPE} & \textbf{SQA} & \textbf{VQA$^{\text{Text}}$} & \textbf{AI2D} & \textbf{GQA} & \textbf{MMB-EN\_en} & \textbf{OCRBench} & \textbf{Avg.} \\
    \midrule
    Original & 1862.17 & 85.88 & 69.51 & 58.21 & 55.21 & 61.97 & 64.09 & 31.30 & 100.0\% \\
    \midrule
    \multicolumn{10}{l}{\textbf{192 tokens}} \\
    SparseVLM & 1787.00 & 85.30 & 68.62 & 57.75 & 54.83 & 59.48 & 63.75 & 30.70 & 98.26\% \\
    VisionZip & 1780.38 & 85.50 & 68.91 & 57.33 & \textbf{55.54} & 59.26 & 63.75 & 31.20 & \underline{98.52}\% \\
    DivPrune & 1771.79 & \textbf{87.22} & 69.87 & 56.61 & 54.92 & 60.07 & 62.46 & 30.40 & 98.18\% \\
    DART & \textbf{1834.58} & 81.65 & 68.96 & 56.90 & 54.66 & 59.52 & 62.88 & 30.00 & 97.45\% \\
    Vispruner & 1795.88 & 86.44 & 68.17 & 57.45 & 54.89 & 59.58 & 62.80 & 30.60 & 98.15\% \\
    PruneSID & 1765.46 & 85.85 & 68.72 & 55.07 & 54.86 & 59.17 & 63.14 & 28.30 & 96.50\% \\
    GRIP+decoder+sl & 1786.38 & 86.00 & 69.11 & \textbf{57.86} & 54.27 & \underline{60.61} & \underline{64.09} & \underline{31.60} & 98.99\% \\
    GRIP+decoder+rl & 1773.87 & \underline{87.05} & \textbf{71.00} & \underline{57.79} & \underline{55.18} & \textbf{61.79} & \textbf{65.89} & \textbf{31.70} & \textbf{100.22}\% \\
    GRIP+encoder+sl & 1781.90 & 85.86 & 68.72 & 57.65 & 54.40 & 59.77 & 62.71 & 30.90 & 98.14\% \\
    GRIP+encoder+rl & 1789.48 & 85.98 & 69.94 & 57.60 & 54.89 & 59.82 & 62.97 & 30.60 & 98.47\% \\
    \midrule
    \multicolumn{10}{l}{\textbf{128 tokens}} \\
    SparseVLM & 1767.36 & 84.90 & 68.57 & 56.71 & 54.47 & 58.38 & 63.66 & 28.10 & 96.48\% \\
    VisionZip & 1761.72 & 83.20 & 68.86 & 56.83 & 54.70 & 57.67 & 62.20 & \textbf{29.80} & 96.57\% \\
    DivPrune & 1713.90 & \textbf{86.94} & 69.75 & 55.89 & 54.24 & 59.27 & 62.29 & 28.80 & \underline{96.59}\% \\
    DART & \textbf{1803.67} & 77.73 & 69.36 & 55.85 & 53.89 & 57.78 & 61.60 & 27.50 & 94.74\% \\
    Vispruner & 1751.49 & 85.05 & 69.11 & 56.80 & 54.50 & 58.29 & 61.60 & 29.00 & 96.45\% \\
    PruneSID & 1733.12 & 84.32 & 67.97 & 54.45 & 54.53 & 57.91 & 61.94 & 27.50 & 94.91\% \\
    GRIP+decoder+sl & 1769.43 & 85.37 & 68.82 & 56.87 & 53.98 & 59.27 & 63.32 & 28.20 & 96.68\% \\
    GRIP+decoder+rl & 1751 & \underline{86.58} & \textbf{71.15} & \textbf{57.17} & \textbf{55.57} & \textbf{60.57} & \textbf{65.46} & 28.1 & \textbf{98.22}\% \\
    GRIP+encoder+sl & 1770.90 & 84.45 & 68.57 & 57.00 & 54.86 & 58.54 & 62.37 & 29.80 & 97.04\% \\
    GRIP+encoder+rl & 1771.85 & 83.99 & 69.87 & 56.93 & 54.83 & 58.66 & 62.46 & 29.50 & 97.12\% \\
    \midrule
    \multicolumn{10}{l}{\textbf{64 tokens}} \\
    SparseVLM & 1600.00 & 77.40 & 69.71 & 53.44 & 53.14 & 53.68 & 59.88 & 21.20 & 89.02\% \\
    VisionZip & 1687.62 & 77.00 & 68.96 & 55.49 & 54.53 & 55.10 & 60.22 & 28.20 & 93.32\% \\
    DivPrune & 1623.98 & \textbf{85.65} & 69.42 & 54.61 & 53.63 & \textbf{57.62} & 59.45 & 27.10 & 93.76\% \\
    DART & 1532.20 & 59.50 & 68.86 & 50.39 & 51.68 & 51.14 & 54.81 & 21.40 & 83.40\% \\
    Vispruner & 1651.82 & 82.74 & 69.11 & \textbf{55.72} & 54.57 & 55.76 & 60.05 & 27.90 & \underline{93.98}\% \\
    PruneSID & 1669.87 & 84.85 & 68.52 & 54.67 & 53.89 & 57.10 & 59.54 & 27.40 & 93.90\% \\
    GRIP+decoder+sl  & 1583.82 & 79.86 & 69.36 & 54.51 & 53.72 & 54.48 & 59.62 & 21.20 & 89.68\% \\
    GRIP+decoder+rl & 1582.40 & 80.60 & 70.22 & 54.99 & 53.79 & 54.40 & 59.79 & 22.80 & 90.71\% \\
    GRIP+encoder+sl & \textbf{1690.10} & 77.91 & 68.52 & 54.54 & 53.98 & 56.50 & 61.60 & \textbf{28.60} & 93.77\% \\
    GRIP+encoder+rl & 1686.77 & 78.77 & \textbf{69.77} & 54.79 & \textbf{54.65} & 56.79 & \textbf{61.69} & 28.00 & \textbf{94.14}\% \\
    \bottomrule
    \end{tabular}
    }
\vspace{-1em}
\end{table*}

At 192 tokens, our \textit{decoder+rl+sparse} variant achieves a Ratio\_all of 100.22, surpassing all baselines and even slightly exceeding the uncompressed original on several benchmarks.
At the more aggressive 64-token budget, where most baselines degrade substantially, our method maintains competitive performance, demonstrating the robustness of the RL-based optimization strategy.

\subsection{Accuracy--Efficiency Trade-off}

\begin{figure}[t]
  \centering
  \includegraphics[width=0.95\linewidth]{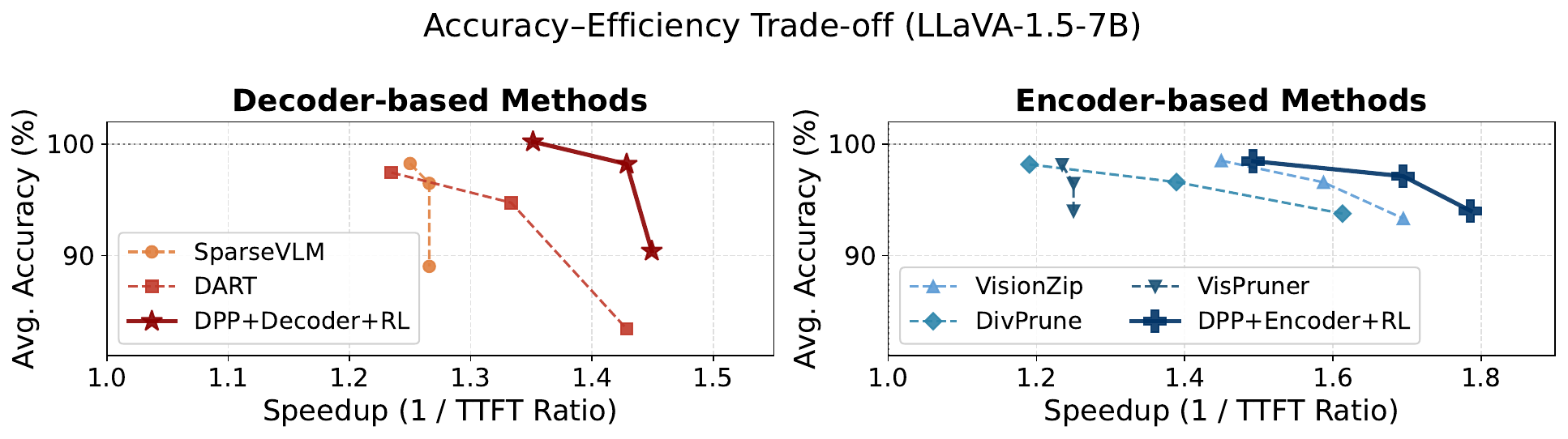}
  \caption{Accuracy--efficiency trade-off on LLaVA-1.5-7B across token budgets (64, 128, 192). GRIP-VLM achieves the best Pareto frontier among all methods in both encoder and decoder configurations.}
  \label{fig:tradeoff}
\end{figure}

As shown in Figure~\ref{fig:tradeoff}, GRIP-VLM achieves the optimal Pareto frontier across both encoder and decoder configurations.
At equal accuracy, our decoder-based variant achieves \textbf{15\%} faster inference than the strongest baseline, while the encoder-based variant reaches the highest overall speedup of $1.79\times$ at 64 tokens.

\subsection{Ablation Study}

Table~\ref{tab:ablation_study} reports the contribution of two key design choices: the budget-aware modulator and the hybrid reward function.


\begin{table*}[t]
\vspace{-1em}
    \centering
    \caption{Ablation study of different module variants under various token budgets.}
    \resizebox{\textwidth}{!}{
    \begin{tabular}{l ccc ccc ccc c}
        \toprule
        \textbf{Methods} & \textbf{MME} & \textbf{POPE} & \textbf{SQA} & \textbf{TextVQA} & \textbf{ai2d} & \textbf{GQA} & \textbf{MMBench} & \textbf{OCRBench} & \textbf{Ratio\_all (\%)} & $\Delta$ \\
        \midrule
        Original & 1862.17 & 85.88 & 69.51 & 58.21 & 55.21 & 61.97 & 64.09 & 31.30 & - & - \\
        \midrule
        \multicolumn{11}{l}{\textbf{128 tokens}} \\
        \midrule
        Original & 1751.00 & 86.58 & 71.15 & 57.17 & 55.57 & 60.57 & 65.46 & 28.10 & 98.22 & - \\
        Decoder w/o modulator & 1753.69 & 86.24 & 68.96 & 57.08 & 54.83 & 60.35 & 64.43 & 28.60 & 97.56 & -0.72 \\
        Decoder w/o mixed reward & 1758.00 & 85.96 & 68.96 & 57.11 & 53.89 & 59.49 & 63.40 & 29.20 & 97.21 & -1.07 \\
        \midrule
        \multicolumn{11}{l}{\textbf{64 tokens}} \\
        \midrule
        Original & 1582.40 & 80.60 & 70.22 & 54.99 & 53.79 & 54.40 & 59.79 & 22.80 & 90.71 & - \\
        Decoder w/o modulator & 1510.80 & 80.59 & 69.96 & 54.67 & 53.72 & 54.48 & 59.62 & 21.20 & 89.44 & -1.26 \\
        Decoder w/o mixed reward & 1566.11 & 82.14 & 69.86 & 54.86 & 53.72 & 55.31 & 58.42 & 21.90 & 90.27 & -0.44 \\
        \bottomrule
    \end{tabular}
    }
    \vspace{2mm} 
    
    \label{tab:ablation_study}
\vspace{-1em}
\end{table*}

\paragraph{Effect of the budget-aware modulator.}
Removing the modulator and instead deploying a model trained at 192 tokens directly to smaller budgets (128 and 64 tokens) leads to consistent degradation across all benchmarks.
At 64 tokens the gap widens to $-1.26$ points in Ratio\_all, confirming that the modulator's ability to perceive and adapt to the current pruning pressure is essential for cross-budget generalization.
Without it, the scorer operates under a distribution mismatch between training and inference compression ratios, causing it to retain suboptimal token subsets.

\paragraph{Effect of the hybrid reward function.}
Replacing the hybrid reward with a uniform $-\mathcal{L}$ reward degrades Ratio\_all by $-1.07$ points at 128 tokens.
This validates the core motivation of our reward design: for verifiable tasks , a binary correctness signal provides a sharper and more faithful learning target than a continuous loss proxy, which can be gamed through logit recalibration without improving actual prediction correctness.
The sigmoid-normalized reward for open-ended tasks further stabilizes training by anchoring the reward scale to the Stage~I SFT baseline, ensuring that the RL agent receives a positive signal only when it genuinely surpasses the supervised initialization.

\subsection{Generalization to Other Models}

To verify that GRIP-VLM is not tailored to a specific backbone, we evaluate Decoder+RL on LLaVA-1.5-13B and compare it against SparseVLM at matched token budgets (176, 110, and 55 tokens).
Table~\ref{tab:performance_comparison} shows that our method consistently outperforms SparseVLM across all budgets and benchmarks.
At the most aggressive 55-token setting, Decoder+RL improves Ratio\_all by $+3.09$ points (90.45\% vs.\ 87.36\%), which mirrors the trend observed on the 7B backbone and confirms that the RL-based scoring strategy transfers across model scales without any architecture-specific tuning.


\begin{table*}[t]
    \centering
    \caption{Generalization to LLaVA-1.5-13B. GRIP-VLM (Decoder+RL) consistently outperforms the SparseVLM baseline across all token budgets (176, 110, and 55 tokens), demonstrating that our method generalizes beyond the 7B backbone used in the main experiments.}
    \resizebox{\textwidth}{!}{
    \begin{tabular}{lccccccccc}
    \toprule
    \textbf{Model / Config} & \textbf{MME} & \textbf{POPE} & \textbf{SQA} & \textbf{TEXT\_VQA} & \textbf{AI2D} & \textbf{GQA} & \textbf{MMBench} & \textbf{OCRBench} & \textbf{Ratio\_all (\%)} \\
    \midrule
    llava-13b & 1826.00 & 85.99 & 72.78 & 61.19 & 59.26 & 63.25 & 68.90 & 33.60 & - \\
    \midrule
    \multicolumn{10}{l}{\textbf{176 tokens}} \\
    SparseVLM(176) & 1813.32 & 85.17 & 73.33 & 59.48 & 57.38 & 59.95 & 68.30 & 32.90 & 98.12 \\
    Decoder+RL(176) & 1829.38 & 85.60 & 73.98 & 59.57 & 58.76 & 60.28 & 68.69 & 33.40 & \textbf{99.04} \\
    \midrule
    \multicolumn{10}{l}{\textbf{110 tokens}} \\
    SparseVLM(110) & 1797.59 & 84.28 & 74.07 & 59.07 & 57.29 & 58.66 & 67.70 & 29.60 & 96.32 \\
    Decoder+RL(110) & 1786.86 & 85.56 & 72.73 & 59.76 & 57.58 & 59.72 & 68.38 & 32.30 & \textbf{97.74} \\
    \midrule
    \multicolumn{10}{l}{\textbf{55 tokens}} \\
    SparseVLM(55) & 1652.62 & 76.22 & 72.48 & 55.86 & 56.74 & 55.59 & 63.57 & 17.80 & 87.36 \\
    Decoder+RL(55) & 1731.92 & 80.19 & 72.97 & 56.88 & 56.54 & 56.14 & 65.24 & 21.30 & \textbf{90.45} \\
    \bottomrule
    \end{tabular}
    }
    
    \label{tab:performance_comparison}
    \vspace{-1em}
    \end{table*}

\subsection{Effectiveness Analysis}

\begin{figure}[t]
  \centering
  \begin{minipage}[t]{0.38\linewidth}
    \centering
    \includegraphics[width=\linewidth]{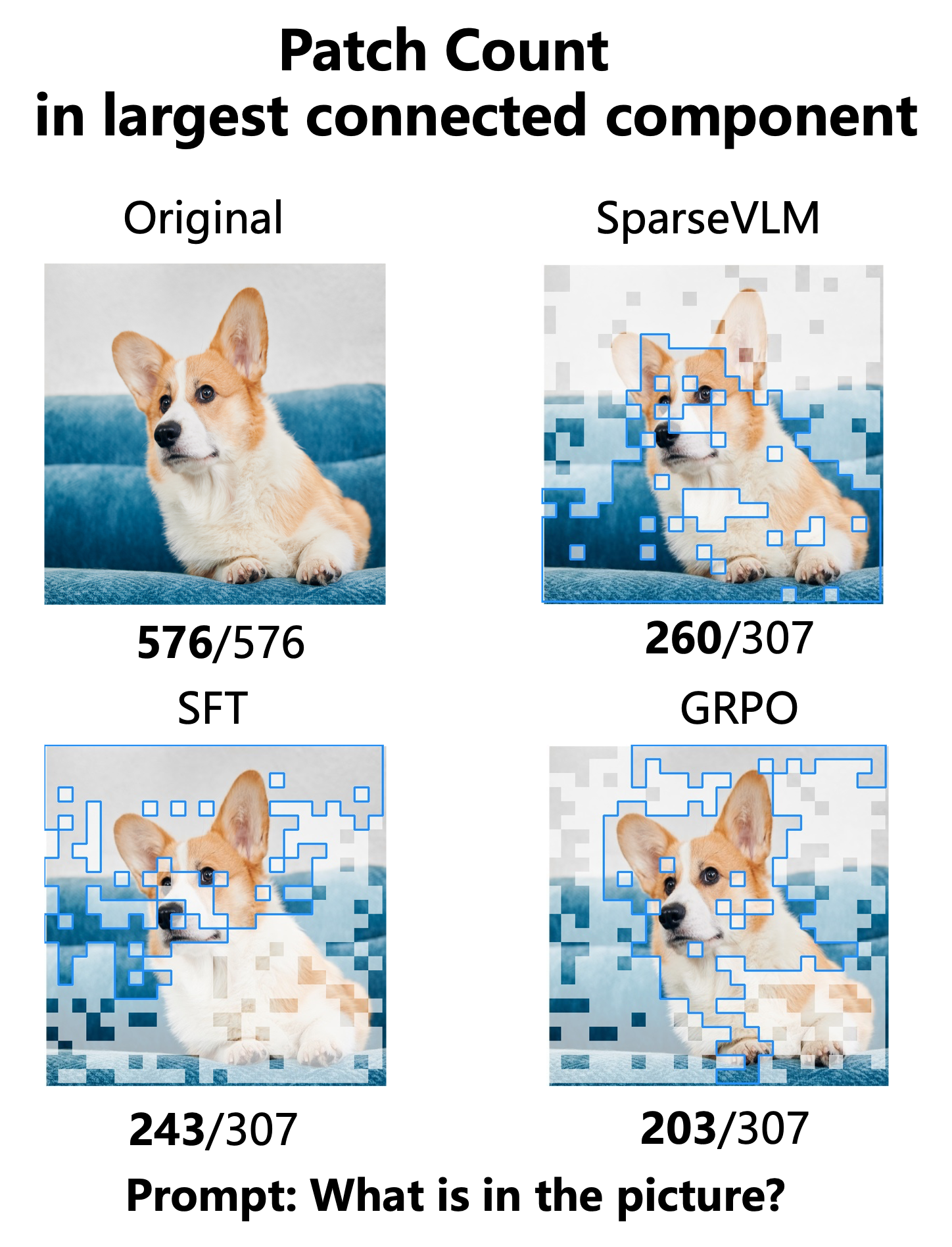}
  \end{minipage}
  \hfill
  \begin{minipage}[t]{0.42\linewidth}
    \centering
    \includegraphics[width=\linewidth]{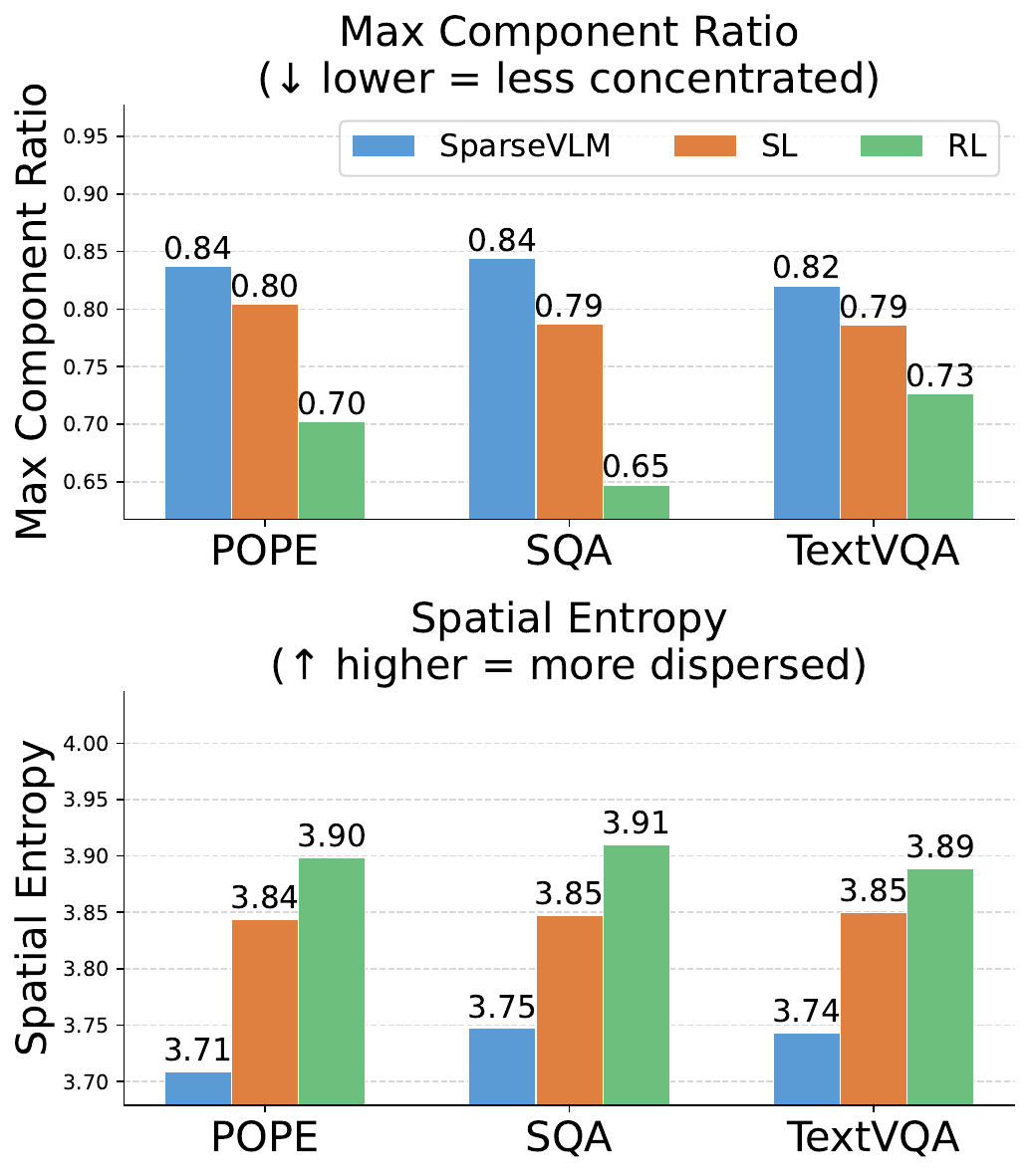}
  \end{minipage}
  \caption{\textbf{Left:} Qualitative visualization of token retention patterns under different pruning methods. RL-based pruning selects finer-grained, spatially dispersed tokens compared to heuristic and SL methods, which tend to retain large contiguous blobs.
  \textbf{Right:} Pruning granularity analysis across POPE, SQA, and TextVQA benchmarks, measured by Max Component Ratio and Spatial Entropy (see Appendix~\ref{app:granularity_metrics} for definitions).}
  \label{fig:granularity}
\end{figure}

A key insight into GRIP-VLM's superiority is its fine-grained token selection (Figure~\ref{fig:granularity}).
Visual attention in VLMs inherently forms spatially contiguous clusters. Consequently, heuristic methods tend to retain redundant, contiguous token ``blobs'' rather than discriminating within them. Supervised learning (SL) methods suffer similarly; their continuous relaxations bias the scorer toward spatially coherent regions to maintain smooth gradients.

Our RL-based approach breaks this coarse-grained bias. By directly optimizing the task reward, the RL agent explores the discrete combinatorial space without being anchored to smooth-gradient solutions. This allows it to discover \emph{scattered but highly informative} tokens that are crucial for reasoning yet isolated from dominant attention clusters.
Quantitatively, GRIP-VLM consistently achieves the lowest max-component ratio and highest spatial entropy. This confirms that our fine-grained, non-redundant retention preserves a more diverse set of visual cues, directly boosting multi-modal accuracy (see qualitative cases in Appendix~\ref{app:case_study}).

\section{Conclusion}
The core insight of this work is that token pruning is a combinatorial problem with non-monotone, non-Lipschitz structure that gradient-based methods cannot handle,and that RL's stochastic sampling naturally provides the token-level credit assignment needed to navigate it.
GRIP-VLM validates this insight empirically across diverse benchmarks.
Future work includes extending to video and high-resolution inputs, and improving the sample efficiency of the RL exploration stage.

\bibliographystyle{unsrt}
\bibliography{main}

@String(CVPR  = {IEEE Conf. Comput. Vis. Pattern Recog.})

@String(CVPR  = {CVPR})

@String(CVPR= {IEEE Conf. Comput. Vis. Pattern Recog.})

@article{zhu2023minigpt,
  title={Minigpt-4: Enhancing vision-language understanding with advanced large language models},
  author={Zhu, Deyao and Chen, Jun and Shen, Xiaoqian and Li, Xiang and Elhoseiny, Mohamed},
  journal={arXiv preprint arXiv:2304.10592},
  year={2023}
}

@article{wang2024qwen2,
  title={Qwen2-vl: Enhancing vision-language model's perception of the world at any resolution},
  author={Wang, Peng and Bai, Shuai and Tan, Sinan and Wang, Shijie and Fan, Zhihao and Bai, Jinze and Chen, Keqin and Liu, Xuejing and Wang, Jialin and Ge, Wenbin and others},
  journal={arXiv preprint arXiv:2409.12191},
  year={2024}
}

@article{bai2025qwen2,
  title={Qwen2. 5-vl technical report},
  author={Bai, Shuai and Chen, Keqin and Liu, Xuejing and Wang, Jialin and Ge, Wenbin and Song, Sibo and Dang, Kai and Wang, Peng and Wang, Shijie and Tang, Jun and others},
  journal={arXiv preprint arXiv:2502.13923},
  year={2025}
}

@article{li2024llava,
  title={Llava-onevision: Easy visual task transfer},
  author={Li, Bo and Zhang, Yuanhan and Guo, Dong and Zhang, Renrui and Li, Feng and Zhang, Hao and Zhang, Kaichen and Zhang, Peiyuan and Li, Yanwei and Liu, Ziwei and others},
  journal={arXiv preprint arXiv:2408.03326},
  year={2024}
}

@misc{liu2024llavanext,
    title={LLaVA-NeXT: Improved reasoning, OCR, and world knowledge},
    url={https://llava-vl.github.io/blog/2024-01-30-llava-next/},
    author={Liu, Haotian and Li, Chunyuan and Li, Yuheng and Li, Bo and Zhang, Yuanhan and Shen, Sheng and Lee, Yong Jae},
    month={January},
    year={2024}
}

@article{vaswani2017attention,
  title={Attention is all you need},
  author={Vaswani, Ashish and Shazeer, Noam and Parmar, Niki and Uszkoreit, Jakob and Jones, Llion and Gomez, Aidan N and Kaiser, {\L}ukasz and Polosukhin, Illia},
  journal={Advances in neural information processing systems},
  volume={30},
  year={2017}
}

@inproceedings{li2023blip,
  title={Blip-2: Bootstrapping language-image pre-training with frozen image encoders and large language models},
  author={Li, Junnan and Li, Dongxu and Savarese, Silvio and Hoi, Steven},
  booktitle={International conference on machine learning},
  pages={19730--19742},
  year={2023},
  organization={PMLR}
}

@misc{instructblip,
      title={InstructBLIP: Towards General-purpose Vision-Language Models with Instruction Tuning}, 
      author={Wenliang Dai and Junnan Li and Dongxu Li and Anthony Meng Huat Tiong and Junqi Zhao and Weisheng Wang and Boyang Li and Pascale Fung and Steven Hoi},
      year={2023},
      eprint={2305.06500},
      archivePrefix={arXiv},
      primaryClass={cs.CV}
}

@article{fu2023mme,
  title={Mme: A comprehensive evaluation benchmark for multimodal large language models},
  author={Fu, Chaoyou and Chen, Peixian and Shen, Yunhang and Qin, Yulei and Zhang, Mengdan and Lin, Xu and Yang, Jinrui and Zheng, Xiawu and Li, Ke and Sun, Xing and others},
  journal={arXiv preprint arXiv:2306.13394},
  year={2023}
}

@article{liu2023mmbench,
  title={Mmbench: Is your multi-modal model an all-around player?},
  author={Liu, Yuan and Duan, Haodong and Zhang, Yuanhan and Li, Bo and Zhang, Songyang and Zhao, Wangbo and Yuan, Yike and Wang, Jiaqi and He, Conghui and Liu, Ziwei and others},
  journal={arXiv preprint arXiv:2307.06281},
  year={2023}
}

@inproceedings{hudson2019gqa,
  title={Gqa: A new dataset for real-world visual reasoning and compositional question answering},
  author={Hudson, Drew A and Manning, Christopher D},
  booktitle={CVPR},
  year={2019}
}

@article{liu2024ocrbench,
  title={OCRBench: on the hidden mystery of OCR in large multimodal models},
  author={Liu, Yuliang and Li, Zhang and Huang, Mingxin and Yang, Biao and Yu, Wenwen and Li, Chunyuan and Yin, Xu-Cheng and Liu, Cheng-Lin and Jin, Lianwen and Bai, Xiang},
  journal={Science China Information Sciences},
  volume={67},
  number={12},
  pages={220102},
  year={2024},
  publisher={Springer}
}

@article{touvron2023llama,
  title={Llama 2: Open foundation and fine-tuned chat models},
  author={Touvron, Hugo and Martin, Louis and Stone, Kevin and Albert, Peter and Almahairi, Amjad and Babaei, Yasmine and Bashlykov, Nikolay and Batra, Soumya and Bhargava, Prajjwal and Bhosale, Shruti and others},
  journal={arXiv preprint arXiv:2307.09288},
  year={2023}
}

@article{grattafiori2024llama,
  title={The llama 3 herd of models},
  author={Grattafiori, Aaron and Dubey, Abhimanyu and Jauhri, Abhinav and Pandey, Abhinav and Kadian, Abhishek and Al-Dahle, Ahmad and Letman, Aiesha and Mathur, Akhil and Schelten, Alan and Vaughan, Alex and others},
  journal={arXiv preprint arXiv:2407.21783},
  year={2024}
}

@article{wang2024internvideo2,
  title={Internvideo2: Scaling video foundation models for multimodal video understanding},
  author={Wang, Yi and Li, Kunchang and Li, Xinhao and Yu, Jiashuo and He, Yinan and Chen, Guo and Pei, Baoqi and Zheng, Rongkun and Xu, Jilan and Wang, Zun and others},
  journal={Arxiv e-prints},
  pages={arXiv--2403},
  year={2024}
}

@article{wang2024exploring,
  title={Exploring the reasoning abilities of multimodal large language models (mllms): A comprehensive survey on emerging trends in multimodal reasoning},
  author={Wang, Yiqi and Chen, Wentao and Han, Xiaotian and Lin, Xudong and Zhao, Haiteng and Liu, Yongfei and Zhai, Bohan and Yuan, Jianbo and You, Quanzeng and Yang, Hongxia},
  journal={arXiv preprint arXiv:2401.06805},
  year={2024}
}

@article{bai2023qwen,
  title={Qwen technical report},
  author={Bai, Jinze and Bai, Shuai and Chu, Yunfei and Cui, Zeyu and Dang, Kai and Deng, Xiaodong and Fan, Yang and Ge, Wenbin and Han, Yu and Huang, Fei and others},
  journal={arXiv preprint arXiv:2309.16609},
  year={2023}
}

@article{yang2024qwen2,
  title={Qwen2. 5 technical report},
  author={Yang, An and Yang, Baosong and Zhang, Beichen and Hui, Binyuan and Zheng, Bo and Yu, Bowen and Li, Chengyuan and Liu, Dayiheng and Huang, Fei and Wei, Haoran and others},
  journal={arXiv preprint arXiv:2412.15115},
  year={2024}
}

@article{li2024mini,
  title={Mini-gemini: Mining the potential of multi-modality vision language models},
  author={Li, Yanwei and Zhang, Yuechen and Wang, Chengyao and Zhong, Zhisheng and Chen, Yixin and Chu, Ruihang and Liu, Shaoteng and Jia, Jiaya},
  journal={arXiv:2403.18814},
  year={2024}
}

@article{tang2023video,
  title={Video understanding with large language models: A survey},
  author={Tang, Yunlong and Bi, Jing and Xu, Siting and Song, Luchuan and Liang, Susan and Wang, Teng and Zhang, Daoan and An, Jie and Lin, Jingyang and Zhu, Rongyi and others},
  journal={arXiv preprint arXiv:2312.17432},
  year={2023}
}

@article{liu2024visual,
  title={Visual instruction tuning},
  author={Liu, Haotian and Li, Chunyuan and Wu, Qingyang and Lee, Yong Jae},
  journal={Advances in neural information processing systems},
  year={2024}
}

@inproceedings{liu2024improved,
  title={Improved baselines with visual instruction tuning},
  author={Liu, Haotian and Li, Chunyuan and Li, Yuheng and Lee, Yong Jae},
  booktitle={Proceedings of the IEEE/CVF Conference on Computer Vision and Pattern Recognition},
  pages={26296--26306},
  year={2024}
}

@article{zhang2024beyond,
  title={Beyond training: Dynamic token merging for zero-shot video understanding},
  author={Zhang, Yiming and Zhao, Zhuokai and Chen, Zhaorun and Ding, Zenghui and Yang, Xianjun and Sun, Yining},
  journal={arXiv preprint arXiv:2411.14401},
  year={2024}
}

@article{wang2025data,
  title={Data whisperer: Efficient data selection for task-specific llm fine-tuning via few-shot in-context learning},
  author={Wang, Shaobo and Jin, Xiangqi and Wang, Ziming and Wang, Jize and Zhang, Jiajun and Li, Kaixin and Wen, Zichen and Li, Zhong and He, Conghui and Hu, Xuming and others},
  journal={arXiv preprint arXiv:2505.12212},
  year={2025}
}

@inproceedings{chen2024imagefastv,
  title={An image is worth 1/2 tokens after layer 2: Plug-and-play inference acceleration for large vision-language models},
  author={Chen, Liang and Zhao, Haozhe and Liu, Tianyu and Bai, Shuai and Lin, Junyang and Zhou, Chang and Chang, Baobao},
  booktitle={European Conference on Computer Vision},
  pages={19--35},
  year={2024},
  organization={Springer}
}

@article{shang2024llavaprumerge,
  title={Llava-prumerge: Adaptive token reduction for efficient large multimodal models},
  author={Shang, Yuzhang and Cai, Mu and Xu, Bingxin and Lee, Yong Jae and Yan, Yan},
  journal={arXiv preprint arXiv:2403.15388},
  year={2024}
}

@inproceedings{singh2019textvqa,
  title={Towards vqa models that can read},
  author={Singh, Amanpreet and Natarajan, Vivek and Shah, Meet and Jiang, Yu and Chen, Xinlei and Batra, Dhruv and Parikh, Devi and Rohrbach, Marcus},
  booktitle={Proceedings of the IEEE/CVF conference on computer vision and pattern recognition},
  pages={8317--8326},
  year={2019}
}

@article{lu2022scienceqa,
  title={Learn to explain: Multimodal reasoning via thought chains for science question answering},
  author={Lu, Pan and Mishra, Swaroop and Xia, Tanglin and Qiu, Liang and Chang, Kai-Wei and Zhu, Song-Chun and Tafjord, Oyvind and Clark, Peter and Kalyan, Ashwin},
  journal={Advances in Neural Information Processing Systems},
  volume={35},
  pages={2507--2521},
  year={2022}
}

@inproceedings{kembhavi2016ai2d,
  title={A diagram is worth a dozen images},
  author={Kembhavi, Aniruddha and Salvato, Mike and Kolve, Eric and Seo, Minjoon and Hajishirzi, Hannaneh and Farhadi, Ali},
  booktitle={European conference on computer vision},
  pages={235--251},
  year={2016},
  organization={Springer}
}

@article{li2023pope,
  title={Evaluating object hallucination in large vision-language models},
  author={Li, Yifan and Du, Yifan and Zhou, Kun and Wang, Jinpeng and Zhao, Wayne Xin and Wen, Ji-Rong},
  journal={arXiv preprint arXiv:2305.10355},
  year={2023}
}

@article{liu2023visualllava,
  title={Visual instruction tuning},
  author={Liu, Haotian and Li, Chunyuan and Wu, Qingyang and Lee, Yong Jae},
  journal={Advances in neural information processing systems},
  volume={36},
  pages={34892--34916},
  year={2023}
}

@misc{wang2024smarttrimadaptivetokensattention,
      title={SmartTrim: Adaptive Tokens and Attention Pruning for Efficient Vision-Language Models}, 
      author={Zekun Wang and Jingchang Chen and Wangchunshu Zhou and Haichao Zhu and Jiafeng Liang and Liping Shan and Ming Liu and Dongliang Xu and Qing Yang and Bing Qin},
      year={2024},
      eprint={2305.15033},
      archivePrefix={arXiv},
      primaryClass={cs.CL},
      url={https://arxiv.org/abs/2305.15033}, 
}

@misc{rao2021dynamicvitefficientvisiontransformers,
      title={DynamicViT: Efficient Vision Transformers with Dynamic Token Sparsification}, 
      author={Yongming Rao and Wenliang Zhao and Benlin Liu and Jiwen Lu and Jie Zhou and Cho-Jui Hsieh},
      year={2021},
      eprint={2106.02034},
      archivePrefix={arXiv},
      primaryClass={cs.CV},
      url={https://arxiv.org/abs/2106.02034}, 
}

@misc{wen2025efficientmultimodallargelanguage,
      title={Efficient Multi-modal Large Language Models via Progressive Consistency Distillation}, 
      author={Zichen Wen and Shaobo Wang and Yufa Zhou and Junyuan Zhang and Qintong Zhang and Yifeng Gao and Zhaorun Chen and Bin Wang and Weijia Li and Conghui He and Linfeng Zhang},
      year={2025},
      eprint={2510.00515},
      archivePrefix={arXiv},
      primaryClass={cs.CV},
      url={https://arxiv.org/abs/2510.00515}, 
}

@misc{zhang2025sparsevlm,
      title={SparseVLM: Visual Token Sparsification for Efficient Vision-Language Model Inference}, 
      author={Yuan Zhang and Chun-Kai Fan and Junpeng Ma and Wenzhao Zheng and Tao Huang and Kuan Cheng and Denis Gudovskiy and Tomoyuki Okuno and Yohei Nakata and Kurt Keutzer and Shanghang Zhang},
      year={2025},
      eprint={2410.04417},
      archivePrefix={arXiv},
      primaryClass={cs.CV},
      url={https://arxiv.org/abs/2410.04417}, 
}

@misc{bai2025qwen25vltechnicalreport,
      title={Qwen2.5-VL Technical Report}, 
      author={Shuai Bai and Keqin Chen and Xuejing Liu and Jialin Wang and Wenbin Ge and Sibo Song and Kai Dang and Peng Wang and Shijie Wang and Jun Tang and Humen Zhong and Yuanzhi Zhu and Mingkun Yang and Zhaohai Li and Jianqiang Wan and Pengfei Wang and Wei Ding and Zheren Fu and Yiheng Xu and Jiabo Ye and Xi Zhang and Tianbao Xie and Zesen Cheng and Hang Zhang and Zhibo Yang and Haiyang Xu and Junyang Lin},
      year={2025},
      eprint={2502.13923},
      archivePrefix={arXiv},
      primaryClass={cs.CV},
      url={https://arxiv.org/abs/2502.13923}, 
}

@misc{bolya2023tokenmergingvitfaster,
      title={Token Merging: Your ViT But Faster}, 
      author={Daniel Bolya and Cheng-Yang Fu and Xiaoliang Dai and Peizhao Zhang and Christoph Feichtenhofer and Judy Hoffman},
      year={2023},
      eprint={2210.09461},
      archivePrefix={arXiv},
      primaryClass={cs.CV},
      url={https://arxiv.org/abs/2210.09461}, 
}

@misc{fu2025framefusioncombiningsimilarityimportance,
      title={FrameFusion: Combining Similarity and Importance for Video Token Reduction on Large Vision Language Models}, 
      author={Tianyu Fu and Tengxuan Liu and Qinghao Han and Guohao Dai and Shengen Yan and Huazhong Yang and Xuefei Ning and Yu Wang},
      year={2025},
      eprint={2501.01986},
      archivePrefix={arXiv},
      primaryClass={cs.CV},
      url={https://arxiv.org/abs/2501.01986}, 
}

@misc{zhu2025visionselectorendtoendlearnablevisual,
      title={VisionSelector: End-to-End Learnable Visual Token Compression for Efficient Multimodal LLMs}, 
      author={Jiaying Zhu and Yurui Zhu and Xin Lu and Wenrui Yan and Dong Li and Kunlin Liu and Xueyang Fu and Zheng-Jun Zha},
      year={2025},
      eprint={2510.16598},
      archivePrefix={arXiv},
      primaryClass={cs.CV},
      url={https://arxiv.org/abs/2510.16598}, 
}

@misc{perez2017filmvisualreasoninggeneral,
      title={FiLM: Visual Reasoning with a General Conditioning Layer}, 
      author={Ethan Perez and Florian Strub and Harm de Vries and Vincent Dumoulin and Aaron Courville},
      year={2017},
      eprint={1709.07871},
      archivePrefix={arXiv},
      primaryClass={cs.CV},
      url={https://arxiv.org/abs/1709.07871}, 
}

@misc{chen2025sftrlearlyinvestigation,
      title={SFT or RL? An Early Investigation into Training R1-Like Reasoning Large Vision-Language Models}, 
      author={Hardy Chen and Haoqin Tu and Fali Wang and Hui Liu and Xianfeng Tang and Xinya Du and Yuyin Zhou and Cihang Xie},
      year={2025},
      eprint={2504.11468},
      archivePrefix={arXiv},
      primaryClass={cs.CL},
      url={https://arxiv.org/abs/2504.11468}, 
}

@misc{dosovitskiy2021imageworth16x16words,
      title={An Image is Worth 16x16 Words: Transformers for Image Recognition at Scale}, 
      author={Alexey Dosovitskiy and Lucas Beyer and Alexander Kolesnikov and Dirk Weissenborn and Xiaohua Zhai and Thomas Unterthiner and Mostafa Dehghani and Matthias Minderer and Georg Heigold and Sylvain Gelly and Jakob Uszkoreit and Neil Houlsby},
      year={2021},
      eprint={2010.11929},
      archivePrefix={arXiv},
      primaryClass={cs.CV},
      url={https://arxiv.org/abs/2010.11929}, 
}

\newpage
\appendix

\section{Full Motivation Results}
\label{app:full_motivation}

\begin{figure*}[h]
  \centering
  \includegraphics[width=0.55\linewidth]{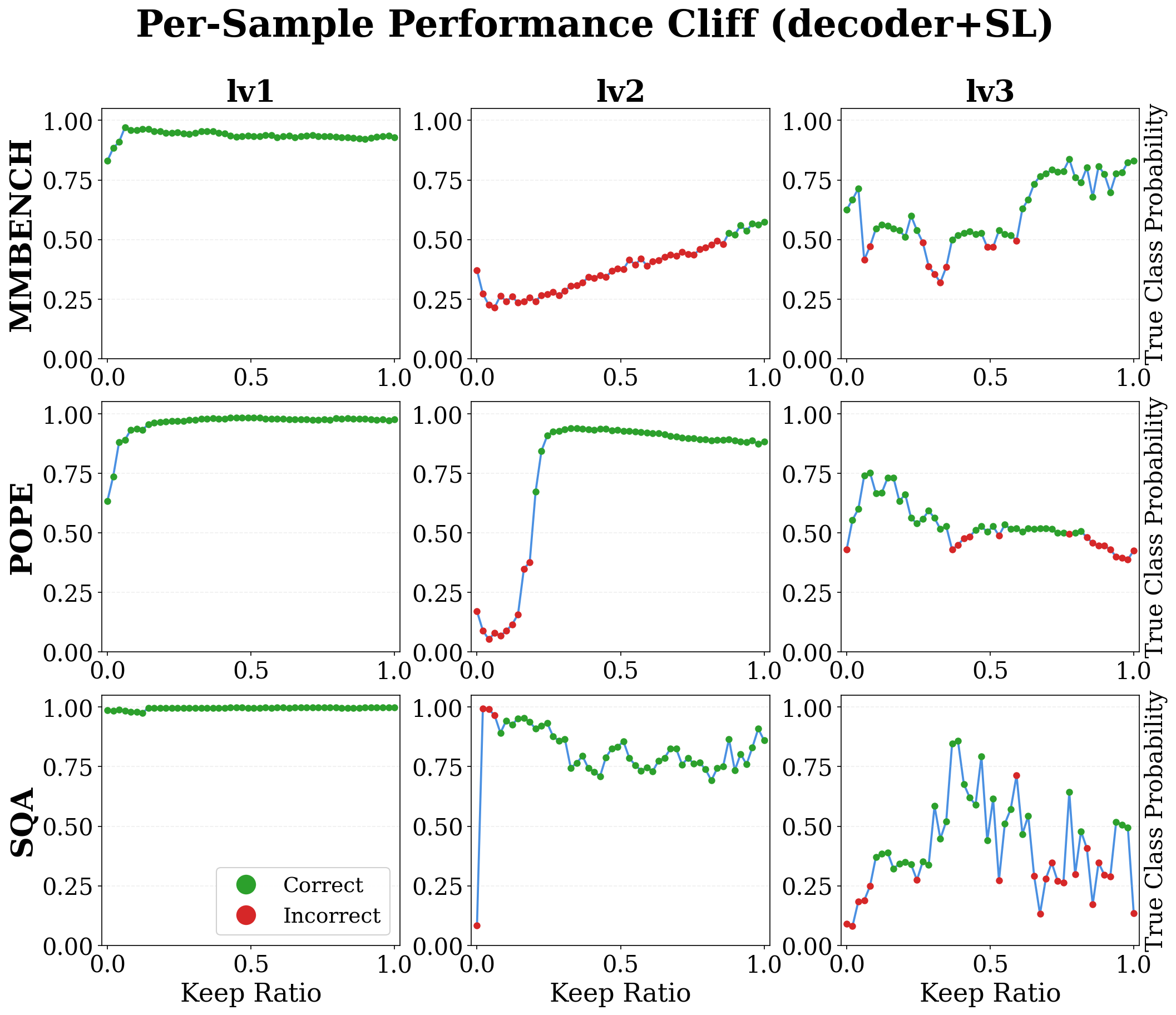}
  \caption{\textbf{Per-Sample Performance Cliff.} Each curve shows the true class probability as the keep ratio varies for representative samples across MMBench, POPE, and SQA under the decoder+SL scorer; green/red dots indicate correct/incorrect predictions.}
  \label{fig:cliff_full}
\end{figure*}

\begin{figure*}[h]
   \vspace{-1em}
  \centering
  \begin{minipage}[t]{0.48\linewidth}
    \centering
    \includegraphics[width=\linewidth]{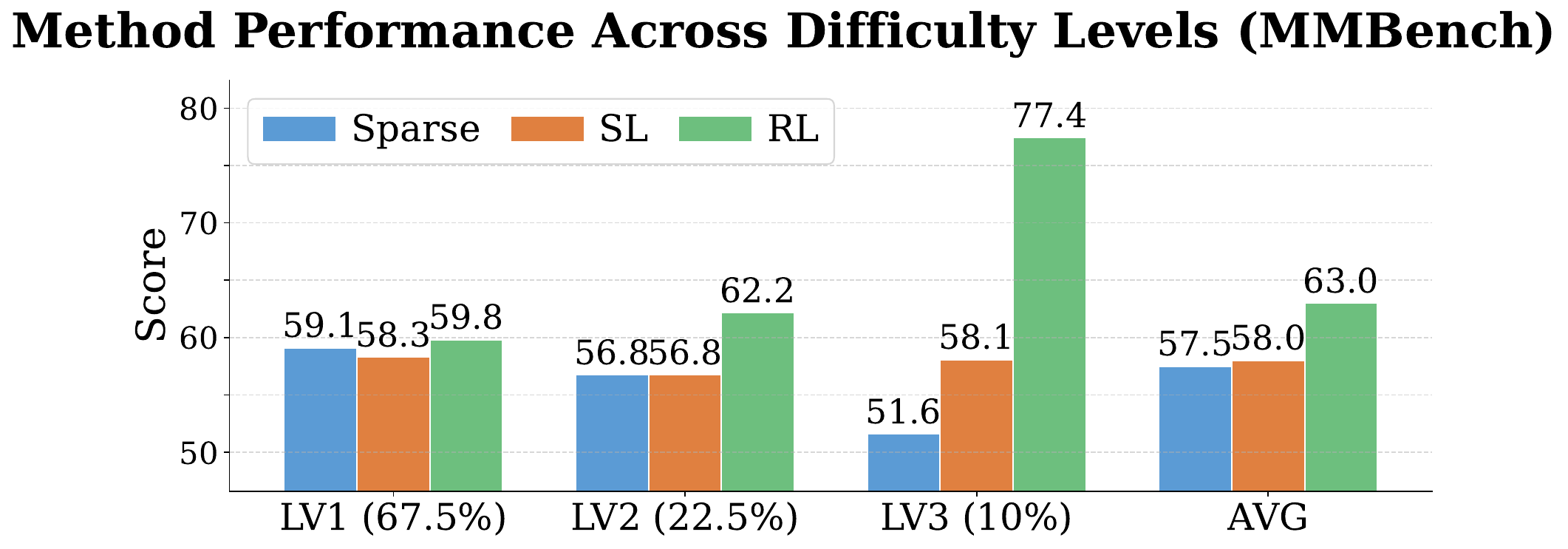}
  \end{minipage}
  \hfill
  \begin{minipage}[t]{0.48\linewidth}
    \centering
    \includegraphics[width=\linewidth]{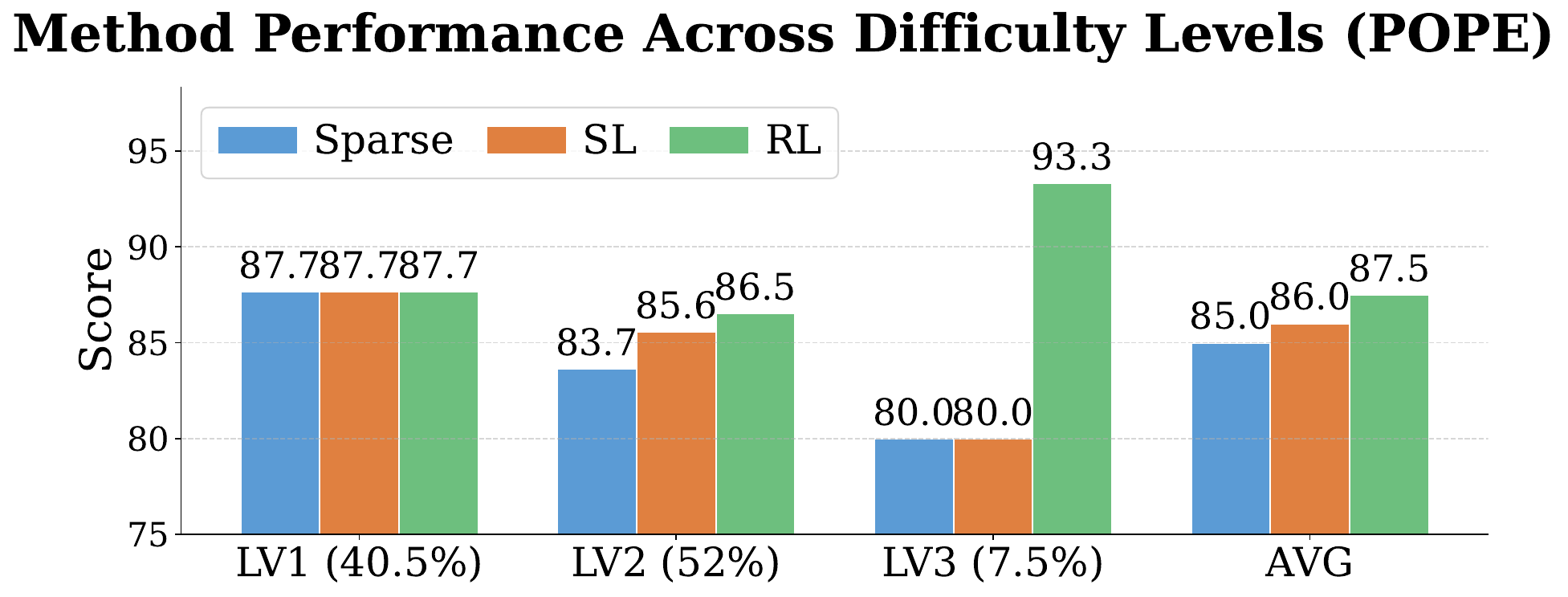}
  \end{minipage}
  \caption{\textbf{Method Performance Across Difficulty Levels.} Accuracy of sparse (heuristic), SL, and RL across difficulty levels (LV1--LV3) and overall average on MMBench (left) and POPE (right).}
  \label{fig:cliff_and_levels_full}
  \vspace{-1em}
\end{figure*}

\section{Case Study: Pruning Granularity}
\label{app:case_study}

\begin{figure}[h]
  \centering
  \includegraphics[width=0.6\linewidth]{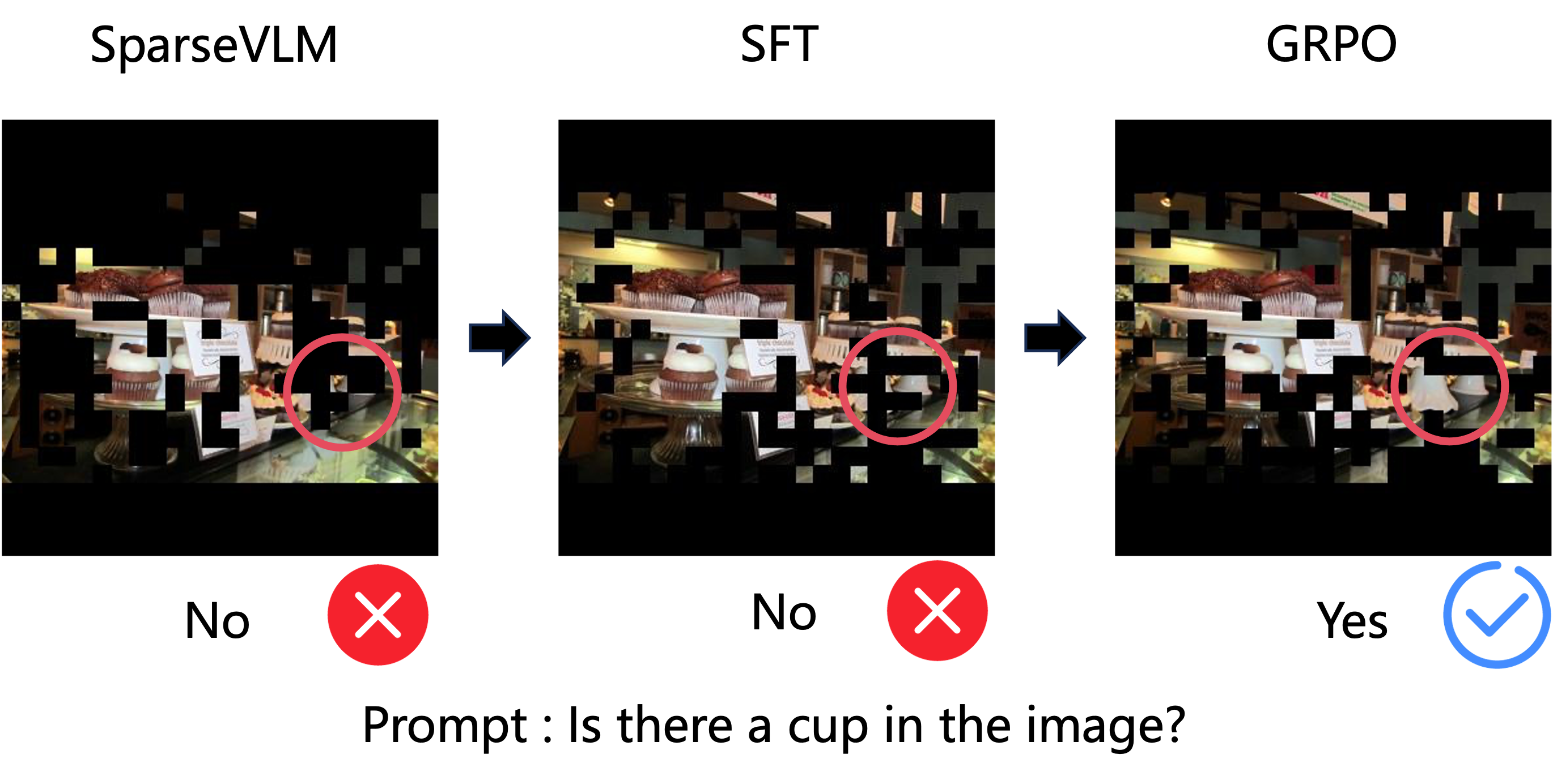}
  \caption{Qualitative case study illustrating the effect of pruning granularity on task correctness.
  Each panel shows the retained tokens (black = pruned) at the same keep ratio under heuristic (left), SL (middle), and RL (right) pruning.
  The red circle highlights the task-critical region (the price label on the cupcake).
  Heuristic and SL methods retain large contiguous blobs and discard the critical token, leading to incorrect predictions ($\times$).
  RL selects a finer-grained, spatially dispersed token set that preserves the critical region, yielding the correct answer (\checkmark).
  This qualitative result corroborates the quantitative granularity analysis in Figure~\ref{fig:granularity}: finer pruning granularity directly translates to higher task accuracy.}
  \label{fig:case_study}
\end{figure}

\section{Training Pipeline}
\label{app:train_pipeline}

\begin{figure}[h]
  \centering
  \includegraphics[width=0.9\linewidth]{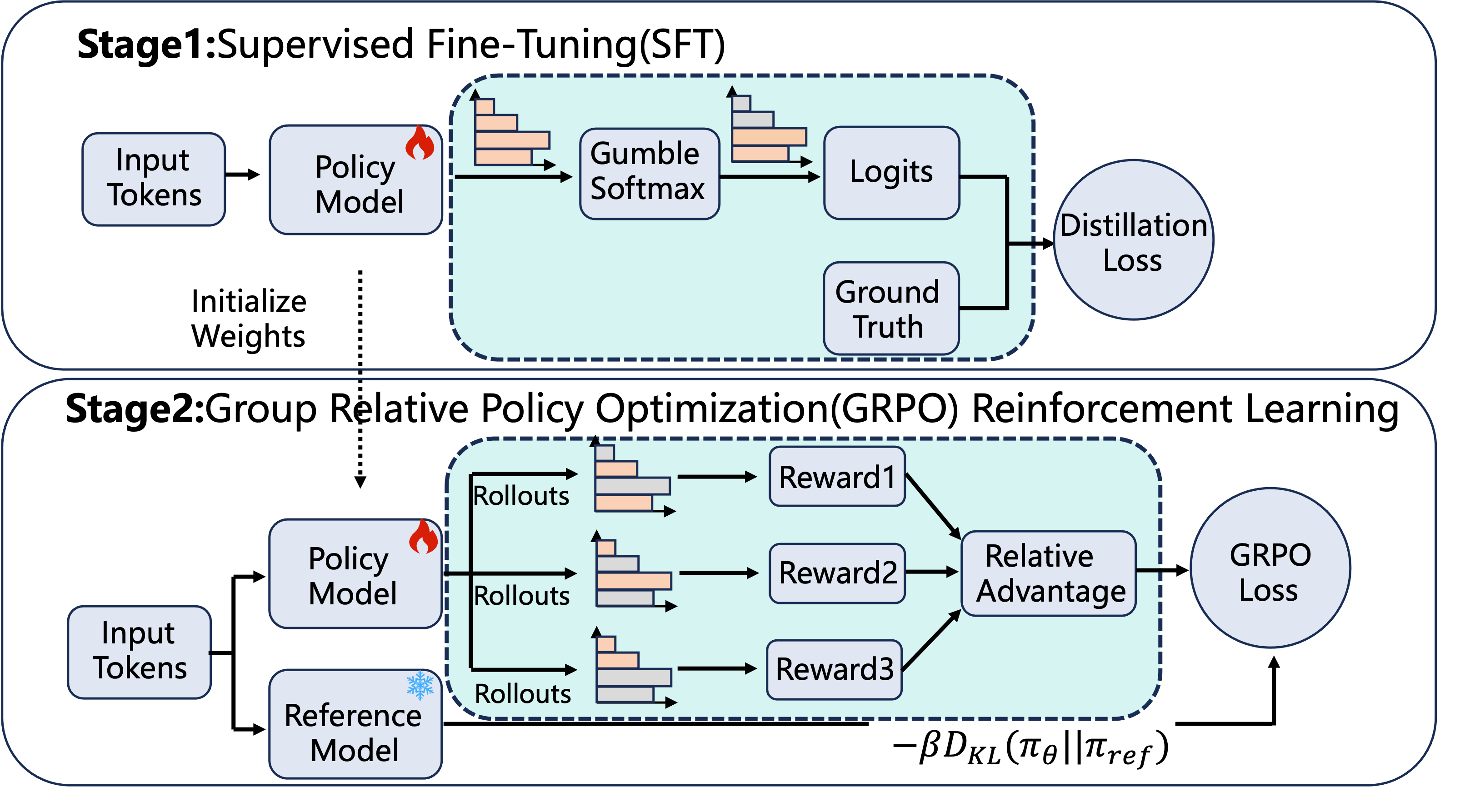}
  \caption{Overview of the two-stage training pipeline of GRIP-VLM. Stage~I performs supervised distillation to warm up the Adaptive Token Scorer using a teacher-student framework. Stage~II freezes the backbone and optimizes the Scorer as an RL policy via GRPO, exploring the discrete token selection space to escape local optima unreachable by gradient-based methods.}
  \label{fig:traing_pipeline}
\end{figure}

\section{Granularity Metrics}
\label{app:granularity_metrics}

We define two metrics used to quantify the spatial granularity of retained token sets in Figure~\ref{fig:granularity}.

\noindent\textbf{Max Component Ratio.} Given the binary retention mask over the spatial token grid, we identify all connected components (4-connectivity) among retained tokens. The max component ratio is defined as the fraction of retained tokens occupied by the single largest connected component. A high value indicates coarse-grained, blob-like retention; lower is better.

\noindent\textbf{Spatial Entropy.} We divide the token grid into a uniform $M \times M$ spatial partition and compute the distribution of retained tokens across cells. Spatial entropy is the Shannon entropy of this distribution. A high value indicates that retained tokens are spread across diverse spatial regions, reflecting fine-grained, scattered selection; higher is better.

\section{Latency Analysis (GQA Dataset)}
\label{app:latency}

Table~\ref{tab:ttft_results} reports the Time to First Token (TTFT) for various methods under different token budgets on the GQA dataset. GRIP-VLM consistently outperforms both the original model and the majority of the baseline methods across all budget settings. Notably, due to its substantial scoring overhead, PruneSID actually incurs a higher latency than the original model (relative ratio $> 1.0$). In contrast, our proposed method effectively reduces the TTFT to $0.56\times$ of the original baseline when operating at a budget of 64 tokens.
\begin{table}[htbp]
    \centering
    \caption{Experimental results of Time to First Token (TTFT) under different token budgets on the GQA dataset.}
    \label{tab:ttft_results}
    \begin{tabular}{lrr}
    \toprule
    \textbf{Method} & \textbf{TTFT (s)} & \textbf{Relative Ratio} \\
    \midrule
    Original & 64.60 & 1.00 \\
    \midrule
    \multicolumn{3}{l}{\textbf{192 tokens}} \\
    SparseVLM & 51.40 & 0.80 \\
    VisionZip & 44.70 & 0.69 \\
    DivPrune & 54.20 & 0.84 \\
    DART & 52.30 & 0.81 \\
    Vispruner & 52.30 & 0.81 \\
    PruneSID & 92.70 & 1.43 \\
    Decoder+RL & 47.50 & 0.74 \\
    Encoder+RL & 43.20 & 0.67 \\
    \midrule
    \multicolumn{3}{l}{\textbf{128 tokens}} \\
    SparseVLM & 50.80 & 0.79 \\
    VisionZip & 40.50 & 0.63 \\
    DivPrune & 46.20 & 0.72 \\
    DART & 48.60 & 0.75 \\
    Vispruner & 51.60 & 0.80 \\
    PruneSID & 89.00 & 1.38 \\
    Decoder+RL & 45.20 & 0.70 \\
    Encoder+RL & 38.40 & 0.59 \\
    \midrule
    \multicolumn{3}{l}{\textbf{64 tokens}} \\
    SparseVLM & 51.10 & 0.79 \\
    VisionZip & 38.40 & 0.59 \\
    DivPrune & 40.30 & 0.62 \\
    DART & 45.20 & 0.70 \\
    Vispruner & 52.00 & 0.80 \\
    PruneSID & 46.70 & 0.72 \\
    Decoder+RL & 44.60 & 0.69 \\
    Encoder+RL & 36.40 & 0.56 \\
    \bottomrule
    \end{tabular}
\end{table}

\section{Benchmark Descriptions}
\label{app:benchmarks}

We evaluate GRIP-VLM on eight widely-used multimodal benchmarks spanning perception, reasoning, and text understanding. Table~\ref{tab:benchmarks} summarizes their key characteristics.

\begin{table}[h]
\centering
\caption{Overview of the eight evaluation benchmarks used in this work.}
\label{tab:benchmarks}
\begin{tabularx}{\textwidth}{@{}>{\raggedright\arraybackslash}p{2.1cm}>{\raggedright\arraybackslash}p{2.5cm}>{\raggedright\arraybackslash}X>{\raggedright\arraybackslash}p{2.0cm}@{}}
\toprule
\textbf{Benchmark} & \textbf{Task Type} & \textbf{Focus} & \textbf{Format} \\
\midrule
MME~\cite{fu2023mme} & Perception \& Cognition & 14 subtasks covering object recognition, counting, scene understanding, commonsense and arithmetic reasoning & Yes/No \\
POPE~\cite{li2023pope} & Hallucination Probing & Object hallucination via three negative sampling strategies (random, popular, adversarial) & Yes/No \\
SQA~\cite{lu2022scienceqa} & Science QA & $\sim$21K K-12 multi-subject science questions with chain-of-thought annotations & Multiple Choice \\
TextVQA~\cite{singh2019textvqa} & Text Recognition & Reading and reasoning about text embedded in natural images (signs, labels, storefronts) & Open-ended \\
AI2D~\cite{kembhavi2016ai2d} & Diagram Understanding & $\sim$5K grade-school scientific diagrams with dense relational annotations & Multiple Choice \\
GQA~\cite{hudson2019gqa} & Compositional Reasoning & 22M scene graph-derived questions targeting multi-step visual reasoning with reduced language bias & Open-ended \\
MMBench~\cite{liu2023mmbench} & Comprehensive Ability & 20 fine-grained ability dimensions evaluated via CircularEval with ChatGPT-based answer matching & Multiple Choice \\
OCRBench~\cite{liu2024ocrbench} & OCR \& Document Understanding & 29 aggregated OCR datasets covering scene text, document QA, and handwriting recognition & Mixed \\
\bottomrule
\end{tabularx}
\end{table}

\noindent\textbf{MME}~\cite{fu2023mme} evaluates VLMs across 14 perception and cognition subtasks using binary yes/no questions, avoiding prompt-sensitivity bias. Perception tasks include object existence, counting, color, and position recognition; cognition tasks cover commonsense reasoning, arithmetic, and code reasoning.

\noindent\textbf{POPE}~\cite{li2023pope} targets object hallucination by probing whether specific objects are present in an image. Three negative sampling strategies of increasing difficulty (random, popular, adversarial) stress-test model reliability and calibration.

\noindent\textbf{ScienceQA (SQA)}~\cite{lu2022scienceqa} is a multimodal multiple-choice benchmark of $\sim$21K questions spanning natural science, social science, and language science at the K-12 level, annotated with chain-of-thought explanations for both answer accuracy and reasoning evaluation.

\noindent\textbf{TextVQA}~\cite{singh2019textvqa} requires models to read and reason about text present within natural images. With 45,336 questions over 28,408 images, it demands tight integration of OCR-level text recognition and visual understanding.

\noindent\textbf{AI2D}~\cite{kembhavi2016ai2d} tests understanding of scientific diagrams (biology, earth science, etc.) through multiple-choice QA. The $\sim$5K diagrams are sourced from grade-school science materials and annotated with dense object segmentations and diagrammatic relations.

\noindent\textbf{GQA}~\cite{hudson2019gqa} focuses on compositional and multi-step visual reasoning over real-world images. Its 22M scene graph-derived questions are designed to reduce the language bias prevalent in earlier VQA datasets, requiring genuine visual grounding.

\noindent\textbf{MMBench}~\cite{liu2023mmbench} provides a systematically structured evaluation across 20 fine-grained ability dimensions, including attribute recognition, spatial reasoning, and social reasoning. It uses a CircularEval strategy with ChatGPT-based answer matching for robust, reproducible scoring.

\noindent\textbf{OCRBench}~\cite{liu2024ocrbench} aggregates 29 OCR-related datasets into a unified suite covering text recognition, scene text QA, document understanding, and handwriting. It is currently the most comprehensive OCR evaluation benchmark for multimodal large language models.

\end{document}